\documentclass{article}

\usepackage{arxiv}
\usepackage{tcolorbox}
\usepackage[utf8]{inputenc} 
\usepackage[T1]{fontenc}    
\usepackage{hyperref}       
\usepackage{url}       
\usepackage{listings}
\usepackage{booktabs}       
\usepackage{amsfonts}       
\usepackage{nicefrac}       
\usepackage{microtype}      
\usepackage{lipsum}		
\usepackage{graphicx}
\usepackage{doi}
\usepackage{algorithm}
\usepackage{algpseudocode}
\usepackage{amsmath}
\usepackage{amssymb}
\usepackage{pifont}

\title{Automating Aggregation Strategy \\ Selection in Federated Learning}

\author{ 
\hspace{0.5mm} Dian S. Y. PANG\textsuperscript{1,2}, Endrias Y. ERGETU\textsuperscript{2}, Eric TOPHAM\textsuperscript{2}, Ahmed E. FETIT\textsuperscript{1}\\\\
	\textsuperscript{1}Department of Computing, Imperial College London, UK\\ \textsuperscript{2} OctaiPipe, London, UK\\
    \\	
}

\renewcommand{\shorttitle}{\textit{Automating aggregation strategy selection in federated learning}}

\hypersetup{
pdftitle={A template for the arxiv style},
pdfsubject={q-bio.NC, q-bio.QM},
pdfauthor={David S.~Hippocampus, Elias D.~Striatum},
pdfkeywords={First keyword, Second keyword, More},
}

\begin{document}
\maketitle
\vspace{-15mm}
\begin{abstract}
Federated Learning enables collaborative model training without centralising data, but its effectiveness varies with the selection of the aggregation strategy. This choice is non-trivial, as performance varies widely across datasets, heterogeneity levels, and compute constraints. We present an end-to-end framework that automates, streamlines, and adapts aggregation strategy selection for federated learning. The framework operates in two modes: a single-trial mode, where large language models infer suitable strategies from user-provided or automatically detected data characteristics, and a multi-trial mode, where a lightweight genetic search efficiently explores alternatives under constrained budgets. Extensive experiments across diverse datasets show that our approach enhances robustness and generalisation under non-IID conditions while reducing the need for manual intervention. Overall, this work advances towards accessible and adaptive federated learning by automating one of its most critical design decisions, the choice of an aggregation strategy.

\end{abstract}

% keywords can be removed

\section{Introduction }
Selecting an appropriate aggregation strategy in Federated Learning (FL) remains a complex and unresolved challenge. Effective decision-making requires expertise across several domains, including machine learning model behaviour, federated optimisation theory, and data characteristics of each edge device. For many practitioners, particularly those without prior experience in FL, this expertise barrier is prohibitive and hampers practical deployment. Even with sufficient expertise, the absence of visibility into client-level data in FL makes the task difficult. Since raw data cannot be centralised in many practical settings, insights into heterogeneity must be inferred indirectly. This is often done through limited metadata or statistical summaries, but such methods are prone to human error and may not reflect the true or evolving state of client datasets, thereby resulting in suboptimal strategies.

Compounding this issue is the context-dependent nature of heterogeneity. While extreme cases such as severely imbalanced label distributions may suggest straightforward strategies, most scenarios involve complex interactions between data distributions and model dynamics, making manual judgement unreliable. Finally, the computational cost of evaluation further limits strategy selection. Testing aggregation methods through full-scale FL training runs requires synchronised communication and computation across clients, rendering exhaustive search infeasible. Conventional AutoML techniques, which rely on repeated trial-and-error, are therefore incompatible with the resource-constrained nature of FL. Motivated by the challenges outlined above, this paper aims to develop an automated approach for selecting FL aggregation strategies that adapts to statistical heterogeneity under different compute constraints. Our goal is to reduce reliance on cumbersome and costly trial-and-error experimentation, thereby enabling accessible deployment of FL in practical settings.

\textbf{Contributions.} In this paper, we present a novel framework for completely automating the selection of aggregation strategy in FL under statistical heterogeneity and different compute constraints. To illustrate our framework, we consider two constraint scenarios: \textit{1) highly constrained single-trial} and \textit{2) lightly constrained multi-trial settings}, reflecting the diverse FL use cases where devices vary widely in their resources. At one extreme, highly constrained edge devices may allow only a single training attempt, while in other cases devices participate in FL primarily for its data locality benefits and face fewer compute limits, permitting several trials. To address highly constrained single-trial settings, we implemented an LLM-based reasoning method that supports both human-in-the-loop prompting and fully automated heterogeneity-driven prompting. For lightly constrained settings, we employed a lightweight genetic search procedure to efficiently refine strategy choices within a bounded budget. To the best of our knowledge, no prior work provides an end-to-end workflow that integrates automated heterogeneity assessment with data-driven selection of aggregation strategies and their parameters.

\section{Related Work}
\label{sec:headings}

\subsection{Data Heterogeneity Challenges in FL}
One significant challenge in federated learning is \textit{data heterogeneity}, also referred to as statistical heterogeneity or non-independent and identically distributed (non-IID) data. In practice, client devices typically generate and store data under diverse conditions, reflecting differences in user behavior, environments, and system usage. As a result, the distribution of local datasets across clients can vary substantially, making the federated optimisation problem much more difficult than the idealised IID setting.

The \textit{de facto} aggregation algorithm, Federated Averaging (FedAvg) \cite{mcmahan2017communication}, updates the global model by computing a weighted average of the local client updates. However, FedAvg implicitly assumes that local client data is IID and that each stochastic gradient update is an unbiased estimate of the global gradient. Under heterogeneous (non-IID) data, this assumption rarely holds. As several works have highlighted \cite{Ye2024HeterogeneousFL, karimireddy2020scaffold, Li2021NonIID}, client devices often overfit to their local objective, which can lead to client drift, where model updates deviate from the direction of the global optimum. Consequently, naive averaging of such updates can result in suboptimal convergence behaviour or slower training. Zhao et al. \cite{Zhao2018NonIID} empirically demonstrate the adverse effect of data heterogeneity on FedAvg, showing a marked reduction in accuracy under non-IID settings, with the performance gap widening as the degree of heterogeneity increases. Moreover, theoretical analyses have established that data heterogeneity increases the number of communication rounds required to achieve a desired accuracy, further exacerbating the communication bottleneck in FL.

Data heterogeneity across clients is typically categorised into four types \cite{Qi2023ModelAggregationSurvey, Ye2024HeterogeneousFL, gao2022surveyheterogeneousfederatedlearning}: (i) \textit{label distribution skew}, where clients have different label proportions; (ii) \textit{feature distribution skew}, where feature distributions differ despite sharing a label space; (iii) \textit{quantity skew}, reflecting imbalanced dataset sizes; and (iv) \textit{concept drift}, where the conditional label distribution varies across clients. This taxonomy is directly relevant to our work, as it establishes the heterogeneity dimensions that ground the design of our adaptive strategy selection framework.

\subsection{Aggregation Strategy Selection in FL}
Several prior works have attempted to assist practitioners in selecting aggregation strategies under non-IID conditions. A common approach is to construct empirical mappings between heterogeneity levels and recommended strategies. Li et al. simulate diverse non-IID settings and evaluate aggregation strategies under these conditions; their empirical findings effectively form mappings between heterogeneity severity and strategy performance. Similarly, Dubey \cite{dubey2025quantifying} presented an empirical framework that quantifies heterogeneity as a metric and recommends strategies empirically shown to be effective under comparable conditions. The strengths of these works lie in their rigorous empirical evaluation of heterogeneity across datasets; however, they exhibit several limitations. While these works provide valuable empirical insight, they exhibit key limitations. Most critically, most work output only \emph{strategy names} and do not tune or recommend \emph{strategy parameters} (e.g., FedProx’s $\mu$ or Krum’s tolerance $f$), despite these parameters having substantial influence on performance. Furthermore, fixed mapping tables cannot adapt to new aggregation strategy algorithms and may not generalise across datasets with different data characteristics, limiting their practical applicability.

A complementary line of research explores the use of large language models (LLMs) to automate configuration generation within FL pipelines. Mawela et al.~\cite{mawela2025webbasedFL} show that LLMs can generate valid experiment configurations, suggest plausible hyperparameter ranges, and integrate seamlessly with FL system specifications. However, their system assumes a fixed aggregation method (FedAvg) and does not address aggregation strategy selection or parameter tuning. Also, this approach still depends on user intervention to prompt the model precisely and rely on hyperparameter optimisation trials, which can be impractical in resource-constrained FL settings.  Shen et al.~\cite{shen2023llmedgeai} similarly use LLMs to produce FL configurations based on user prompts, focusing on experiment scaffolding rather than optimisation. These works demonstrate the feasibility and utility of LLM-driven automation in FL, but they do not provide mechanisms for selecting or fine-tuning aggregation strategies under heterogeneity.

\subsection{Data Heterogeneity Detection}
Diagnosing data heterogeneity is a core component of our LLM-based FL workflow. Reliable quantification of client-level skew provides the basis for informed aggregation strategy selection, yet systematic detection remains under-explored in the literature. Without such diagnostics, practitioners typically rely on full-scale FL experiments to infer heterogeneity effects, incurring unnecessary computational cost. Integrating an explicit detection stage addresses this gap by characterising the type and severity of skew before aggregation. Our heterogeneity detection module is inspired by insights from \cite{kummaya2025fedhetero} and \cite{dubey2025quantifying}, and incorporates specialised algorithms tailored to FL strategy selection.

The Fed-Hetero framework proposed by Kummaya et al. \cite{kummaya2025fedhetero} addresses three key forms of data heterogeneity in FL: quantity skew, label distribution skew, and image skew. The framework begins by measuring weight divergence across client updates, which serves as an initial indicator of potential skew. When divergence exceeds a predefined threshold, more fine-grained measures are applied. For label distribution skew, the Jensen–Shannon Divergence (JSD) is calculated between local and global class distributions, providing a symmetric and bounded measure of difference. For image skew, visual similarity between samples across clients is assessed, quantifying structural and perceptual variation. Finally, quantity skew is measured directly by evaluating differences in dataset sizes across clients. 

The second work by Dubey \cite{dubey2025quantifying} proposed a multi-modal divergence framework to quantify heterogeneity along three dimensions: label distribution skew, feature (covariate) shift, and model output shift. The first two dimensions build on concepts similar to Fed-Hetero but differ in the actual approach. Label skew is measured via JSD between class histograms, while feature shift is assessed using Wasserstein distance over feature embeddings. The key addition is output heterogeneity, where clients average their softmax prediction distributions and JSD is applied to capture divergence in predictive behaviour across clients. This output-level analysis extends detection capabilities to concept drift and model confidence shifts that are often overlooked.

\subsection{Positioning and Gap Analysis}
Despite substantial progress in understanding and mitigating data heterogeneity, existing works fall short of providing an end-to-end solution for aggregation strategy optimisation. Prior empirical mapping studies \cite{Li2021NonIID, dubey2025quantifying} recommend strategy choices based on controlled simulations but do not perform parameter-level optimisation within those strategies. This limitation is significant because aggregation parameters (e.g., FedProx's $\mu$, Krum's $f$, or aggregation weights in adaptive methods) critically influence performance and are often more impactful than the choice of strategy alone. As a result, existing approaches provide only coarse guidance and cannot adapt to new aggregation algorithms or evolving dataset characteristics.

Similarly, LLM-based FL automation systems \cite{mawela2025webbasedFL, shen2023llmedgeai} facilitate configuration generation but do not address the problem of strategy selection or strategy-parameter tuning. These systems still rely on human prompting and conventional hyperparameter optimisation, which limits their applicability in compute-constrained FL scenarios where repeated trials are infeasible. Heterogeneity detection frameworks such as Fed-Hetero \cite{kummaya2025fedhetero} offer useful diagnostic signals but stop short of linking these diagnostics to any optimisation or decision-making mechanism.

\begin{table}[h]
\small
\centering
\caption{Comparison of related work on strategy selection, heterogeneity detection, and automation.}
\label{tab:related_positioning}
\resizebox{\textwidth}{!}{
\begin{tabular}{lccccc}
\toprule
\textbf{Work} &
\textbf{Heterogeneity Detection} &
\textbf{Strategy Selection} &
\textbf{Aggregation Tuning} &
\textbf{LLM Automation} &
\textbf{Compute-Aware} \\
\midrule
Li et al. (2021) & \checkmark & \checkmark (fixed mapping) &  &  &  \\
Dubey (2025) & \checkmark & \checkmark (fixed mapping) &  &  &  \\
Fed-Hetero (2025) & \checkmark &  &  &  &  \\
Mawela et al. (2025) &  &  &  & \checkmark &  \\
Shen et al. (2023) &  &  &  & \checkmark &  \\
\textbf{Ours} & \checkmark (lightweight) & \checkmark (adaptive) & \checkmark & \checkmark & \checkmark \\
\bottomrule
\end{tabular}
}
\end{table}

\textbf{Gap in Prior Work.} To the best of our knowledge, no existing method provides a unified and lightweight framework that (i) diagnoses client-level heterogeneity, (ii) selects an aggregation strategy accordingly, and (iii) fine-tunes the strategy's internal parameters under compute constraints. Because prior works either omit parameter tuning or depend on repeated hyperparameter trials, they cannot serve as direct baselines for the problem addressed here. Consequently, direct experimental comparison is not feasible and custom benchmarks is used and later discussed. Furthermore, the use of LLMs or evolutionary search, particularly genetic algorithms, for tuning aggregation parameters has not been explored in the literature.

This positioning highlights the distinct contribution of our framework: an end-to-end, resource-efficient pipeline that integrates heterogeneity detection with LLM-driven strategy selection under compute constraints, and lightweight genetic search for parameter refinement.

\section{Overview of the Framework}

In this section, we present the overall design of our proposed framework for automating aggregation strategy selection in FL under statistical heterogeneity. The design explicitly considers differences in computational budgets, which constrain the number of trials that can realistically be run; in practice, such constraints arise because edge devices vary widely in their available compute resources. We implemented our framework on top of the \textit{Flower} \cite{beutel2022flower} stack, leveraging its modular design for server–client coordination, PyTorch integration, and aggregation strategy management. Simulations were executed on a single machine, with each client instantiated as an isolated process. This setup enabled controlled data partitioning to emulate heterogeneous non-IID conditions.

\begin{figure}[H]
\centering
\includegraphics[width=0.8\textwidth]{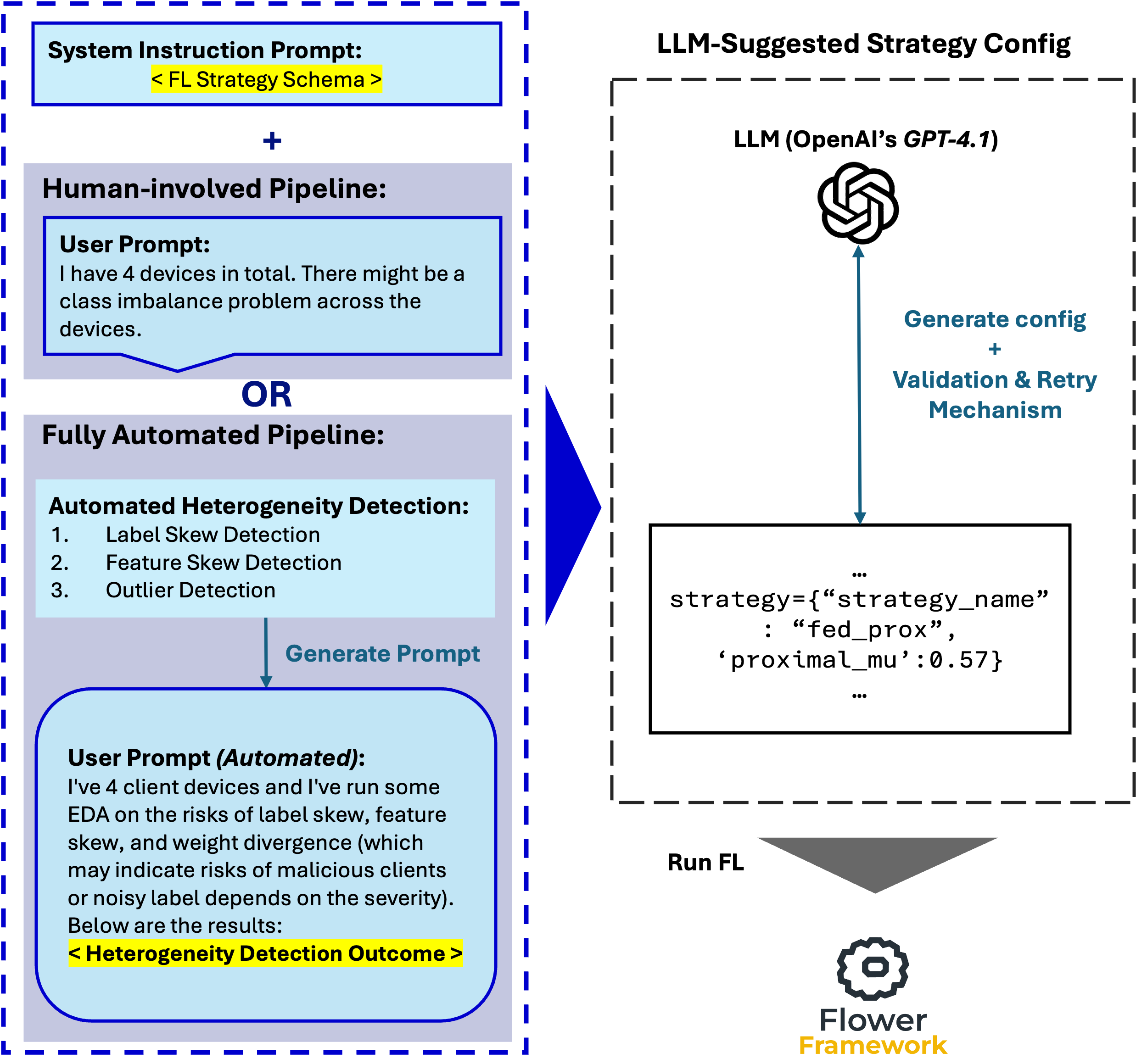}
\caption{Framework design for the single-trial mode. Users can either provide their own description of heterogeneity (human-involved pipeline) or rely on fully automated heterogeneity detection (label skew, feature skew, and outlier detection). In both cases, the information is formatted into an LLM prompt. The LLM generates an aggregation strategy configuration, which is validated before being executed for federated training.}
\label{fig:one_shot_approach}
\end{figure}

In the single-trial mode, our framework must generate a suitable aggregation strategy configuration within one attempt. Two variants of the single-trial mode are proposed as depicted in Figure~\ref{fig:one_shot_approach}, and both variants rely on  OpenAI’s \textit{GPT-4.1}~\cite{openai2025gpt41} LLM to map descriptions of heterogeneity into strategy recommendations but differ in their levels of human involvement. In the \textit{human-involved} variant of the single-trial framework, users provide a textual description of heterogeneity observed across FL clients, such as class imbalance or feature distribution skew. This approach is well-suited to scenarios where expert oversight is desirable, as it ensures transparency and user control over the diagnostic description; the LLM consumes the user-provided input and recommends an appropriate aggregation strategy configuration. A validation and retry mechanism ensures the configuration is executable before it is passed to \textit{Flower} \cite{beutel2022flower}. On the other hand, the \textit{fully automated} variant of the single-trial framework removes the need for human input by automatically diagnosing heterogeneity across clients. Federated analysis is performed to detect label skew, feature skew, and outlier clients, and the diagnostic results are then encoded into a structured prompt for the LLM, which recommends an aggregation strategy in the same way as the human-involved pipeline. This approach is suited to scenarios where human input is unavailable or automation is preferred.

\begin{figure}[h!]
\centering
\includegraphics[width=0.8\textwidth]{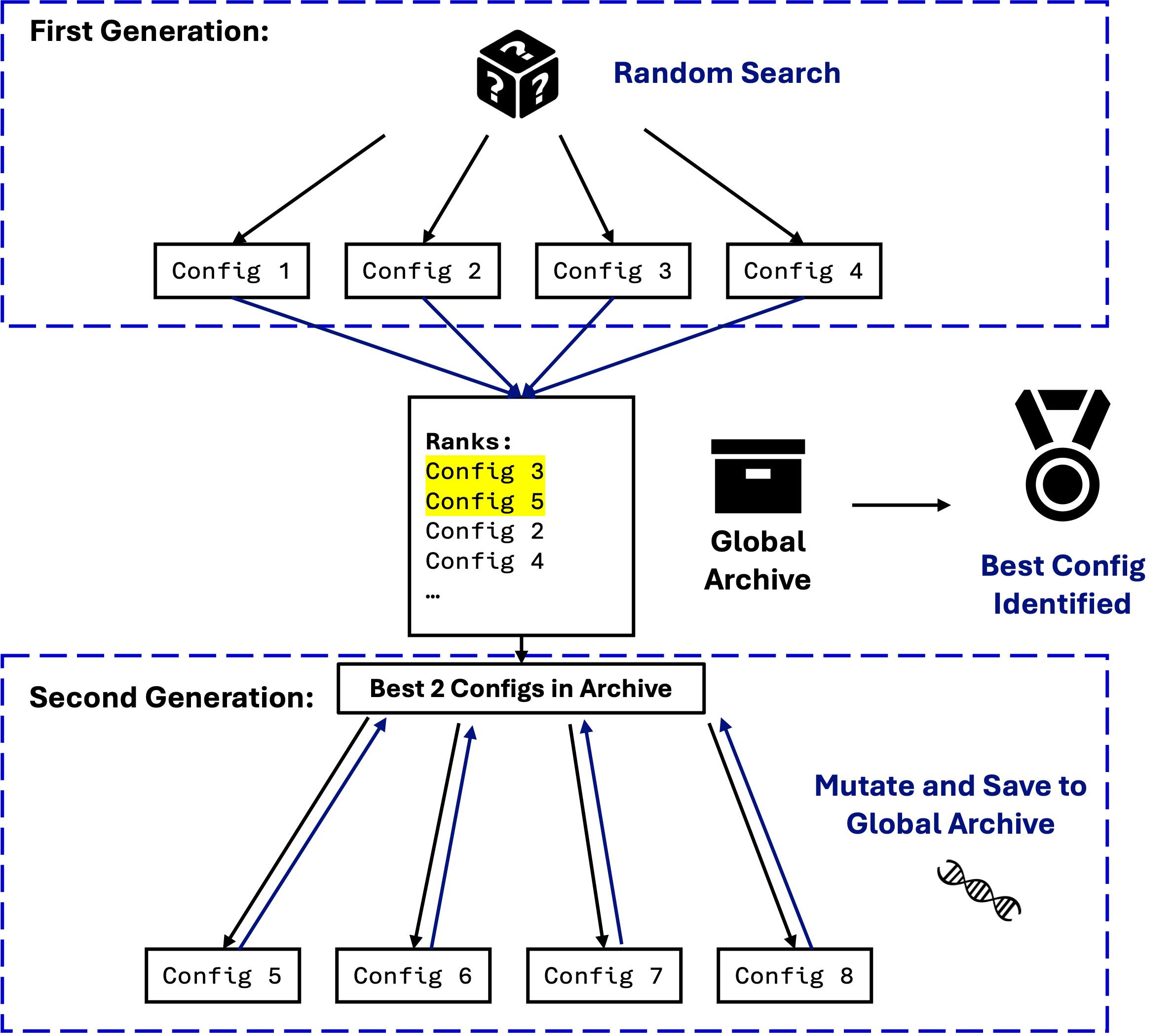}
\caption{Framework design for the multi-trial mode. The first generation of candidates is randomly generated and evaluated. Top-performing configurations are then selected as parents and mutated to form new candidates, which inherit the strategy type but perturb associated parameters. Each candidate is evaluated and stored in a global archive, enabling progressive refinement of aggregation strategies while keeping the search computationally lightweight.}
\label{fig:multi_trial_approach}
\end{figure}

When computational resources are less constrained, strategy optimisation can proceed across multiple trials, hence we also introduce a multi-trial mode. To balance exploration with efficiency, the framework employs a genetic search approach, illustrated in Figure~\ref{fig:multi_trial_approach}. The number of trials is capped at eight, ensuring the method remains lightweight while allowing for parameter tuning across trials. As shown in Figure~\ref{fig:multi_trial_approach}, the search begins with a randomly generated first generation of candidate configurations, which are evaluated and stored in a global archive. From the second generation onward, the top two candidates are selected as “parents”, and new candidates inherit the aggregation strategy type from their parents while perturbing associated parameters within a small local range. Each new configuration is evaluated, ranked, and added to the global archive. This process progressively refines aggregation strategies and parameters, achieving performance comparable to a 50-trial hyperparameter search within a bounded trial budget. The best configuration is finally selected for federated training using \textit{Flower}.

\section{ Development and Evaluation Methodology}
\label{sec:eval}
To arrive at the final design of the framework, development was carried out in four iterative phases, each progressively reducing human intervention while improving optimisation. \textit{Phase 1} established the foundation by using an LLM (OpenAI's \textit{GPT4.1} model~\cite{openai2025gpt41}) to generate syntactically valid and effective aggregation strategies from human-provided descriptions of heterogeneity. \textit{Phase 2} extended this by automating heterogeneity detection, removing the need for manual input. \textit{Phase 3} introduced multi-trial optimisation, exploring both multi-shot prompting and genetic search to refine strategy selection within a bounded number of attempts. Finally, \textit{Phase 4} evaluated each component and the complete framework at scale.

\paragraph{Dataset}
Table~\ref{tab:datasets} summarises the datasets used across all phases. These span tabular, image, text, and reinforcement learning domains, enabling evaluation across diverse data modalities and FL task types.

\begin{table}[h!]
\centering
\small
\caption{Overview of all datasets used in the experimental phases that led to the final design.}
\label{tab:datasets}
\begin{tabular}{|p{3.5cm}|p{2.5cm}|p{3cm}|p{5.5cm}|}
\hline
\textbf{Dataset} & \textbf{Data Type} & \textbf{Task Type} & \textbf{Task Description} \\
\hline
\texttt{NASA Bearing} \cite{citil2023nasa} & Tabular & Fault prediction & Predicting whether a bearing is healthy or faulty from vibration sensor readings. \\
\hline
\texttt{CIFAR-10} \cite{krizhevsky2009learning} & Images & Object \newline recognition & Predicting the object category (e.g., car, dog, airplane) from natural images. \\
\hline
\texttt{Wine Quality} \cite{cortez2009winequality} & Tabular & Quality \newline assessment & Predicting wine quality scores (1-10) based on physicochemical features. \\
\hline
\texttt{Retinal Images} \cite{baptista2023federated, XIA2024117151} & Images & Disease diagnosis & Predicting whether fundus images show glaucomatous disease or healthy eyes.  \\
\hline
\texttt{Sentiment140} \cite{go2009twitter} & Text & Sentiment \newline analysis & Predicting whether a tweet expresses positive or negative sentiment. \\
\hline
\texttt{CartPole} \cite{cartpole} & Reinforcement Learning environment & Control & Predicting optimal actions to balance a pole on a moving cart under varying dynamics. \\
\hline
\end{tabular}
\end{table}

During development, variants of these datasets were created using simulation techniques (see Appendix~\ref{Appendix}). The \texttt{NASA Bearing} dataset served as the primary resource due to its suitability for controlled heterogeneity simulation. Its tabular, binary classification setting enables precise manipulation of data distributions, allowing isolated evaluation of specific heterogeneity types.

We define a set of heterogeneity scenarios based on this dataset, which are used consistently throughout the paper: \textbf{\textit{(i) IID setting}} (no heterogeneity), \textbf{\textit{(ii) label skew}} (imbalanced class distributions across clients), \textbf{\textit{(iii) feature skew}} (distributional shifts via noise), \textbf{\textit{(iv) label poisoning}} (systematic label flipping), \textbf{\textit{(v) noisy labels}} (random label corruption), and \textbf{\textit{(vi) corrupted clients}} (combined feature and label corruption). These scenarios serve as controlled test cases for evaluating detection and optimisation behaviour. Full implementation details are provided in Appendix~\ref{Appendix}.

To improve generalisability, additional datasets were incorporated during development. Partitioned \texttt{CIFAR-10} subsets were used to ensure applicability to image data, particularly under feature skew. Dirichlet partitioning \cite{yurochkin2019bayesian} (see Appendix~\ref{Appendix}) was applied to generate varying degrees of heterogeneity.

Evaluation was then extended to broader settings. The \texttt{Retinal Images} dataset captures real-world heterogeneity arising from differences in clinical data collection. The \texttt{Wine Quality} dataset introduces variation within tabular tasks, extending to multi-class prediction. \texttt{CIFAR-10} was further used to study scalability under varying client device counts and heterogeneity levels. \texttt{Sentiment140} extends the analysis to natural language processing, while the \texttt{CartPole} environment demonstrates applicability to reinforcement learning.

Each dataset was paired with a fixed model representative of its domain (e.g., a feed-forward network for tabular data and a CNN for images). To isolate the effect of aggregation strategy selection, all model architectures and hyperparameters were kept constant across experiments.

\paragraph{Evalation Metrics}
Unless stated otherwise, performance is reported using \textit{mean weighted accuracy} and its \textit{standard deviation} over 10 repeated runs of the same experimental setup. For each run, weighted accuracy is computed across simulated clients over the final five FL rounds, capturing the convergence region.

Weighted accuracy is used as client data in federated settings is typically imbalanced. An unweighted average would overemphasise small or outlier clients, whereas weighting by dataset size better reflects performance across the overall data distribution and aligns with practical deployment scenarios.

\paragraph{Performance Benchmark}
FL performance depends strongly on underlying data distributions, and no universally optimal aggregation strategy exists. To provide robust benchmarks, we define three reference points: \textit{(i) HPO empirical worst}, \textit{(ii) HPO empirical best}, and \textit{(iii) baseline} (FedAvg), where HPO denotes hyperparameter optimisation. FedAvg represents an uninformed default strategy but is often sensitive to heterogeneity, making it an unstable reference point. To establish more reliable bounds, we employ Bayesian optimisation with Optuna \cite{akiba2019optuna}, exploring the strategy space over 50 trials. This yields empirical best and worst performance under reasonable optimisation effort. While computationally expensive, this approach enables rigorous assessment of whether the proposed method outperforms standard baselines and how closely it approaches the empirical optimum.

All runtime measurements were conducted in simulation on a single machine (Apple M3 Pro, 12-core CPU, 18 GB memory) using Python, PyTorch, and the \textit{Flower} framework. Runtime is reported in seconds, with emphasis on relative comparisons rather than absolute values to avoid overemphasising machine-specific performance. Comparisons are made against a typical manual FL run.

\section{Base Infrastructure}
\label{chp6: human_pipeline}

This section presents the foundational infrastructure introduced during \textit{Phase 1}, which enables LLM-based automation of FL strategy selection. This phase produced two key capabilities: \textit{(i)} a scaffold supporting subsequent automation and optimisation, and \textit{(ii)} a human-in-the-loop pipeline for strategy recommendation.

Figure~\ref{fig:one_shot_approach} summarises the base infrastructure, leveraging the \textit{GPT4.1} model~\cite{openai2025gpt41} as the recommendation tool. Central to the design is the system instruction prompt (Template~\ref{instruction_prompt}), which specifies selection requirements and constraints for the LLM. The prompt is invisible to end users but critical for ensuring reliable system integration. A sample user prompt illustrates its interaction with the instruction prompt.

\newcounter{prompt}
\newtcolorbox[auto counter, number within=section, use counter=prompt]{promptbox}[1][]{
  colback=gray!10,
  colframe=black,
  title=System Instruction Prompt~\theprompt,
  #1
}

\newtcolorbox[auto counter, number within=section, use counter=prompt]{promptbox2}[1][]{
  colback=gray!10,
  colframe=black,
  title=Sample User Prompt~\theprompt,
  #1
}
\begin{promptbox}
\label{instruction_prompt}
You are an expert in Federated Learning who helps users decide their FL strategy and associated parameters based on the data heterogeneity they describe.
\\\\
Your ultimate goal is to generate a configuration that orchestrates an FL workflow with optimal predictive performance for the given heterogeneity scenario.

\medskip

\textbf{Return format:} You must return \emph{only} a valid configuration in Python dictionary format.

\medskip

\textbf{Allowed Schema:} \verb|<fl_schema>|

\medskip

Return only a single Python dictionary—no explanations, no extra text.
\\\\
\textbf{Examples of valid output:}

\begin{lstlisting}
{`strategy_name': `fed_avg'}
{`strategy_name': `fed_prox', `proximal_mu': 0.7}
{`strategy_name': `fed_trimmed_avg', `beta': 0.3}
{`strategy_name': `krum', `num_malicious_clients': 1, 
`num_clients_to_keep': 3}
\end{lstlisting}

Your output must be a valid Python dictionary using single quotes.
\end{promptbox}

\begin{promptbox2}
I have 4 client devices and I've run some EDA on the risks of label skew, 
feature skew, and weight divergence which may indicate risks of malicious behaviour.
\label{user_prompt}

\medskip

Below are the results: \\\\
Label Skew: Yes \\
Feature Skew: No \\
Outlier Risk: Yes
\end{promptbox2}
\vspace{1mm}

User inputs are inherently unconstrained, which makes the system prompt essential to steer both content and format. To refine this prompt, we performed iterative tests using the five simulated \texttt{NASA Bearing} datasets, each exhibiting distinct distributional characteristics that encourage the LLM to generate diverse strategy configurations. 

These tests led to three key design choices. Embedding the FL schema proved essential for reliable parameter generation, as it defines valid strategy names and parameter ranges and types. It increased output validity from 2 of 5 to 5 of 5 test cases. Also, few-shot prompting provided valid output examples within the instruction prompt; without it, errors occurred in 2 of 5 cases. Finally, a validation and retry mechanism safeguards against edge cases by checking LLM outputs and allowing up to three retries before raising an informational error.

Prompt format was also carefully evaluated. The first template we explored, \texttt{Format-A}, provides raw distribution statistics, requiring the LLM to interpret numerical data and infer the heterogeneity characteristics before suggesting strategies. \texttt{Format-B} is the chosen user prompt provided above. It incorporates a pre-diagnosed summary with binary flags \textit{yes / no} for label skew, feature skew, and outlier risk. \texttt{Format-C} extends the binary flags by annotating risk severity \textit{high / low / none}.  Table \ref{tab:prompt_comparison} shows that \texttt{Format-B} consistently outperformed the other formats across all simulated heterogeneity settings. The lower performance of \texttt{Format-A} reflects a known LLM limitation, as LLMs are not optimised for mapping raw statistics to FL strategies. \texttt{Format-C} exposes another limitation: insensitivity to severity annotations. In two simulations with severe and moderate label skew, the LLM recommended \texttt{fed\_prox} in both cases but paradoxically assigned larger \texttt{proximal\_mu} values to the moderate scenario, resulting in inappropriate, excessive penalisation. The results identify \texttt{Format-B} as the most effective prompt format.

\begin{table}[h]
    \centering
    \caption{Summary of FL strategy selection performance across prompt formats during \textit{Development Phase 1.}}
    \label{tab:prompt_comparison}
    \begin{tabular}{lccc}
        \toprule
        \textbf{Simulated Heterogeneity (NASA Bearing)} & \textbf{Format-A} & \textbf{Format-B} & \textbf{Format-C} \\
        \midrule
        Label Skew        & 0.6122 $\pm$ 0.0021 & \textbf{0.6360} $\pm$ 0.0058 & 0.6094  $\pm$ 0.0043 \\
        Feature Skew      & 0.7002 $\pm$ 0.0183 & \textbf{0.7022} $\pm$ 0.0241 & 0.6484  $\pm$ 0.0121 \\
        Noisy Label       & 0.6340 $\pm$ 0.0213 & \textbf{0.6548} $\pm$ 0.0410 & 0.6007 $\pm$ 0.0171 \\
        Label Poisoning   & 0.5789 $\pm$ 0.0336 & \textbf{0.6087} $\pm$ 0.0180 & 0.5553 $\pm$ 0.0217 \\
        Corrupted Client  & 0.5626 $\pm$ 0.0149 & \textbf{0.6265} $\pm$ 0.0219 & 0.5144 $\pm$ 0.0172 \\
        IID               & 0.7151 $\pm$ 0.0211 & \textbf{0.7208} $\pm$ 0.0200 & 0.6751 $\pm$ 0.0254 \\
        \bottomrule
    \end{tabular}
\end{table}

The iterative experiments and design choices outlined above achieved full validity on the test cases, laying a solid foundation for subsequent phases. The pipeline established in this section was therefore fixed at the strategy recommendation stage, while the next section will automate the detection step to generate the optimised prompts programmatically.

\section{Automated Heterogeneity Detection}
This section focuses on the features introduced in \textit{Phase 2} of development, which optimised the heterogeneity detection feature of the framework, comprising separate modules for detecting label skew, outliers, and feature skew. Each detection module was tested thoroughly to evaluate whether detection outcomes align with the known heterogeneity.

\subsection{Label Skew Detection Module}
Label skew refers to imbalances in label distributions across clients in a federated learning system. As noted by Zhao et al.~\cite{Zhao2018NonIID}, it is particularly problematic compared with other forms of heterogeneity, as it often slows convergence and degrades predictive accuracy. To detect label skew in our framework, Jensen–Shannon (JS) divergence \cite{kummaya2025fedhetero} is used as it is bounded ($0 \leq JS \leq 1$), scale-invariant, and sensitive to proportional differences, making it an ideal candidate. Each client first computes its empirical label distribution and sends it to the server, which aggregates them into a global distribution. The divergence between each client and the global distribution is then measured using the JS divergence:

\begin{equation}
JS(P \parallel Q) = \frac{1}{2} KL(P \parallel M) + \frac{1}{2} KL(Q \parallel M), \quad M = \frac{1}{2}(P+Q)
\end{equation}

where $P$ is the client distribution, $Q$ is the global distribution, and $M$ is their mixture distribution. A client is flagged as label-skewed if the divergence exceeds a threshold of 0.1. The full procedure is summarised in Algorithm~\ref{algo:label-skew} below.

Two illustrative cases are shown in Figure~\ref{fig:label_skew_demo_binary} and Figure~\ref{fig:label_skew_demo_multi_label}, representing binary and multi-label skewed conditions respectively. In the binary case (\texttt{NASA Bearing}), clients show imbalanced distributions across the two classes. In the multi-label case (\texttt{CIFAR-10} with Dirichlet $\alpha=0.1$~\cite{yurochkin2019bayesian}), Device~2 contains only four labels whereas Device~3 contains ten. As shown in Table~\ref{tab:label_skew_intrinsic}, the JS divergence values correctly reflect these two skewed conditions, and additional tests on IID partition returned near-zero values, rejecting the diagnosis. These observations provided reassurance that JS divergence can reliably capture skew in both binary and multi-label settings.

\begin{algorithm}[h!]
\caption{Label skew detection via Jensen-Shannon divergence.}
\label{algo:label-skew}
\begin{algorithmic}[1]
\State \textbf{Client-side:}
\For{each client $c_i$}
    \State Compute local label distribution $P_i$
    \State Send $P_i$ to the server
\EndFor

\State \textbf{Server-side:}
\State Collect all distributions $\{P_1, P_2, \dots, P_n\}$
\State Align label sets across all clients
\State Compute global distribution: $P_{\text{global}} = \frac{1}{n} \sum_{i=1}^{n} P_i$
\For{each client $c_i$}
    \State Compute $D_{\text{JS}}(P_i \,\|\, P_{\text{global}})$
\EndFor

\State Compute $\max D_{\text{JS}}$ across clients

\If{$\max D_{\text{JS}} > \tau$}
    \State \textbf{Label skew detected}
\Else
    \State \textbf{No significant label skew}

\EndIf
\end{algorithmic}
\end{algorithm}

\begin{figure}[H]
    \centering
    \begin{minipage}{0.49\textwidth}
        \centering
        \includegraphics[width=\textwidth]{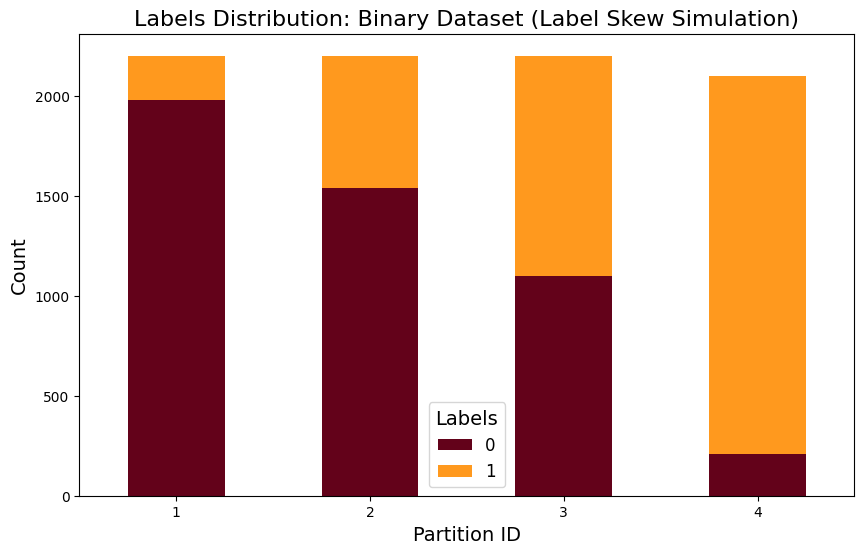}
        \caption{Label skew (binary label).}
        \label{fig:label_skew_demo_binary}
    \end{minipage}
    \hfill
    \begin{minipage}{0.49\textwidth}
        \centering
        \includegraphics[width=\textwidth]{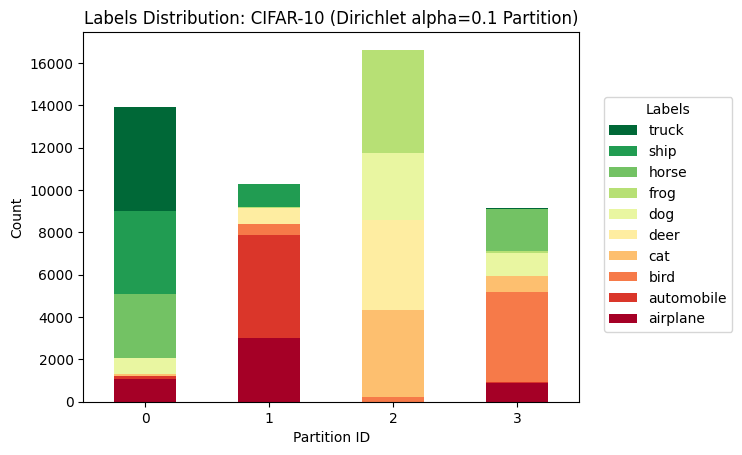}
        \caption{Label skew (multi-label).}
        \label{fig:label_skew_demo_multi_label}
    \end{minipage}
\end{figure}

\begin{table}[h!]
\centering
\begin{tabular}{lccc}
\toprule
\textbf{Simulated Heterogeneity} & \textbf{JS Divergence Value} & \textbf{Diagnosis} \\
\midrule
Skewed \textit{(NASA Bearing, Binary)}     & 0.1890 & Skew \\
Skewed \textit{(CIFAR-10, Multi-label)}    & 0.4341 & Skew \\
IID \textit{(NASA Bearing, Binary)}        & 0.0001 & No Skew \\
IID \textit{(CIFAR-10, Multi-label)}       & 0.0002 & No Skew \\
\bottomrule
\end{tabular}
\caption{Intrinsic evaluation of label skew detection using JS divergence during \textit{Development Phase 2}.}
\label{tab:label_skew_intrinsic}
\end{table}

Note that prior to adopting JS divergence, we investigated the utility of entropy for detecting label skew due to its simplicity. Entropy quantifies uncertainty in a distribution:  

\begin{equation}
H(P) = - \sum_{i=1}^K p_i \log p_i
\end{equation}

where $K$ is the number of labels and $p_i$ is the relative frequency of label $i$. The server aggregated client entropies and defined divergence as the difference between maximum and minimum entropy values. In principle, lower entropy indicates skew, while higher entropy corresponds to balanced distributions. However, we quickly revealed that entropy is unsuitable as it does not generalise across multi-label settings. Specifically, entropy is inherently scale-independent and unbounded; e.g. in a binary classification task ($K=2$) the maximum entropy is $H_{max}=1$, whereas for a 10-class problem ($K=10$) the value of $H_{max} \approx 3.32$, meaning that a single global threshold cannot be applied across datasets. Even when normalised by $H_{max}$, entropy failed to capture the relative frequencies of labels, i.e. two very different distributions can yield similar entropy values. Empirically, the entropy spread for the \texttt{NASA Bearing} above was 0.5007, while for \texttt{CIFAR-10} it produced only 0.1586, despite both being clear examples of skew. This misalignment confirmed that entropy was not a suitable candidate for the label skew detection module.

\subsection{Outlier Detection Module}
In FL, outlier clients are participants whose model updates are significantly different from the others. This form of heterogeneity can arise from adversarial behaviour like model poisoning, corrupted local datasets, or extreme concept drift. If left unaddressed, such clients introduce bias into the aggregated model in proportion to their relative weight in the aggregation process. Early detection of outliers therefore enables mitigation to take place before they affect the global model. In our framework, the following formulation is applied at the second FL training round to detect outliers:  
\begin{tcolorbox}[colback=white, colframe=black, boxrule=0.8pt, arc=4pt]
Let $\mathcal{C} = \{1, 2, \dots, N\}$ be the set of clients. Each client $i$ has parameters
\begin{equation}
W_i = \{ w_i^{(1)}, w_i^{(2)}, \dots, w_i^{(L)} \}.
\end{equation}
\\
The pairwise divergence between clients $i$ and $j$ is defined as
\begin{equation}
D(i,j) = \sum_{\ell=1}^L \, \big\| w_i^{(\ell)} - w_j^{(\ell)} \big\|_2 .
\end{equation}

The divergence score of client $i$ is then
\begin{equation}
\mathrm{Div}(i) = \frac{1}{N-1} \sum_{\substack{j \in \mathcal{C} \\ j \neq i}} D(i,j).
\end{equation}
\end{tcolorbox}

The module initially collects client weights after the second round of federated training using \textit{FedAvg} and ensures that there server computes pairwise divergences between client updates to locate potential outliers. Since FL is inherently stochastic,  randomness in optimisation and sampling can cause divergence values to fluctuate. To avoid false diagnoses, the detection is repeated a small number of times, and outliers are identified by the \textit{frequency} of being flagged. Since this metric does not rely on the absolute divergence values, it generalises better across data modalities. Furthermore, as only two training rounds are run per repetition, the process remains computationally lightweight. A client is finally flagged as an outlier if its divergence value falls within the 90th percentile in at least four repetitions.

\begin{figure}[h!]
    \centering
    \includegraphics[width=0.9\textwidth]{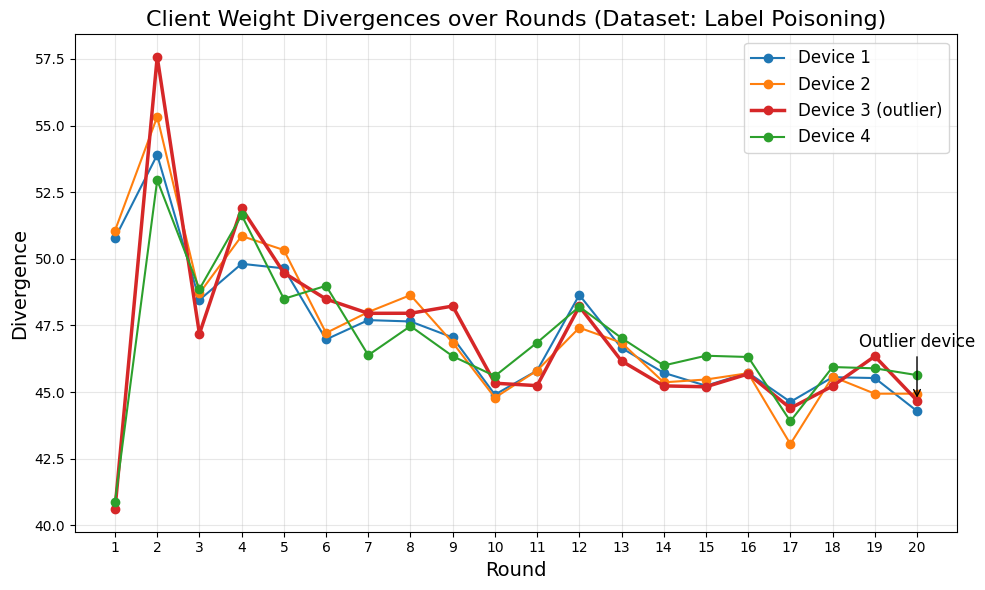}
    \caption{Client weight divergence pattern by FL training rounds in outlier setting.}
    \label{fig:outlier_w_div}
\end{figure}

When outliers are present, divergence values differ most in the earliest training rounds. In \textit{FedAvg}, the global model absorbs local updates and redistributes them, gradually homogenising client weights. As a result, later rounds blur outlier signals, making them harder to detect. The second round provides the clearest signal of raw local differences before aggregation dominates, whereas the first round is excluded due to noise from random initialisation. See Figure~\ref{fig:outlier_w_div} for a demonstration of this behaviour using the simulated label poisoning dataset. Client~3, with all labels flipped, produces updates that diverge markedly from the rest of the clients in the second round, but the strong pattern diminishes over time as aggregation homogenises the models.

When testing the set-up, the module behaved as expected across all heterogeneity scenarios. In the label poisoning and corrupted client datasets, outlier devices were consistently flagged with high frequency (5/5 and 4/5 runs, respectively). By contrast, the noisy label dataset reached a maximum of 3/5 runs. Although relatively high, this falls below the threshold and was therefore not classified as an outlier. This is consistent with expectations, since noisy labels represent a weaker form of distribution shift. For the label skew and feature skew scenarios, detections were random and infrequent (at most 2/5 runs per device), confirming that no true outliers were present.

An alternative that we explored included the method proposed by Kummaya et al. \cite{kummaya2025fedhetero}, where divergence is computed between each client update and the aggregated global model after training. While such approach captures broad differences, two limitations were identified. First, the global model itself may be a biased reference, as aggregation dilutes client-specific signals. Second, the threshold for divergence is dataset-dependent thus difficult to generalise across datasets. Another alternative that we explored was to use a small evaluation dataset that was shared by all clients. In such setup, the server tested each client’s model on the held-out set and flagged those with poor performance as outliers. While effective in principle, this alternative had two major drawbacks. First, it is unrealistic to assume that all clients are willing to share evaluation data, raising privacy concerns. Second, the method is computationally expensive since for each round, the server must run multiple inferences, one for each client’s weights, which is infeasible under FL’s compute constraints. In light of the above, we did not incorporate those alternatives to the design of our framework.

\subsection{Feature Skew Detection Module}
Feature skew arises when feature distributions differ substantially across clients. To detect such skew, we employed a federated principal component analysis (PCA) mechanism. Operating in a reduced-dimensional space mitigates the `curse of dimensionality' and improves robustness to noise; this is valuable in practical settings  where users may not have pre-processed the data with feature selection. In addition, the federated PCA procedure requires only aggregated statistics, ensuring that raw data remain private in transmission. The module operates in two rounds, as detailed below. 

In the first round, each client computes local summary statistics from its dataset. These include per-feature sums, sums of squares, sample counts, and all upper-triangular cross-products required to construct the global covariance matrix. The aggregated statistics are shared without exposing raw data in order preserve privacy. The server then aggregates these values to compute global means, assemble the covariance matrix, and performs PCA to extract the top $k=2$ eigenvectors. 

In the second round, the server broadcasts the global means and the selected principal components to all clients. Each client centres its data using the global means, projects it onto the shared components, and computes the centroid of its projected data in the 2D subspace. These centroids are then returned to the server. To compute divergence, the server compares the client centroids in the shared two-dimensional space by calculating pairwise Euclidean distances between them. If client centroids cluster closely, feature distributions are consistent, whilst large separations indicate skew.
In practice, feature skew is flagged if the maximum pairwise distance exceeds a threshold ($\tau = 1.0$ in this study). 

\begin{figure}[H]
\centering
\includegraphics[width=1\textwidth]{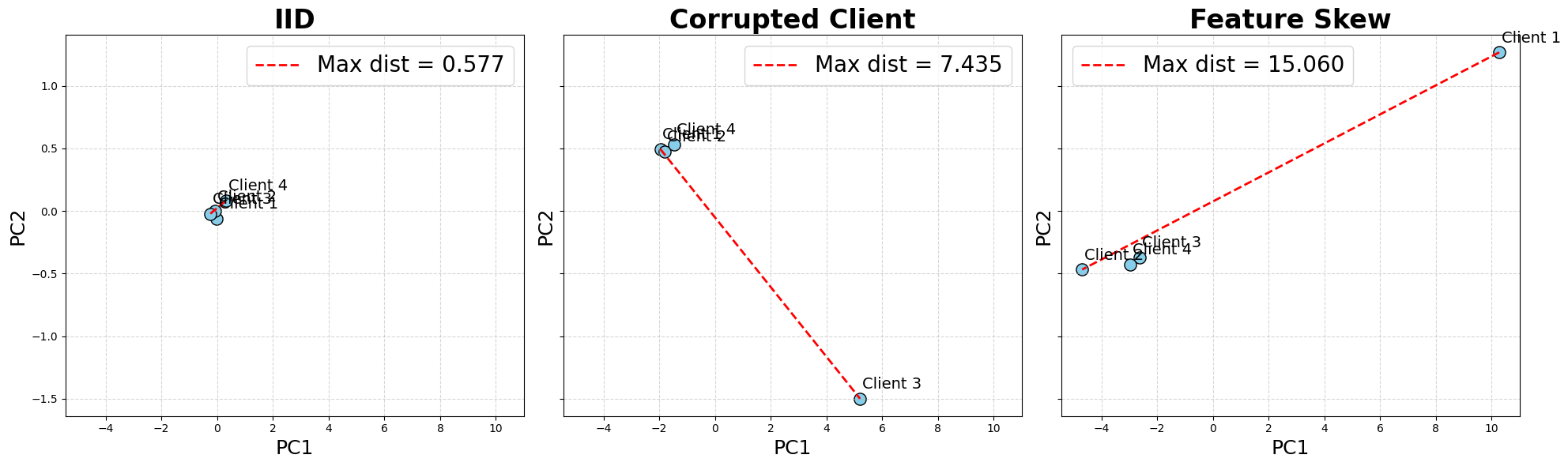}
  \caption{PCA centroid pariwise distances on simulated \texttt{NASA Bearing} datasets.}
  \label{fig:pca_centroid_tabular}
\end{figure}

When tested, the module behaved as expected, with feature skew correctly flagged when present. Figure~\ref{fig:pca_centroid_tabular} illustrates pairwise differences between projected client PCA centroids under three simulated \texttt{NASA Bearing} scenarios. In the IID split, data points were randomly allocated while preserving balanced class ratios. In the corrupted client scenario, one device was perturbed in the feature space. In the feature skew scenario, all clients experienced varying degrees of perturbation, producing inconsistent feature spaces. The results clearly demonstrate that the maximum pairwise distance is negligible in the IID case (0.577), significantly larger in the feature skew scenario (15.060), and intermediate in the corrupted client case (7.435). Notably, in the corrupted client dataset, only Client~3 diverges, while the remaining clients cluster tightly, indicating consistent mappings after PCA projection. Given that the \texttt{NASA Bearing} dataset has 254 features, our tests demonstrate that the federated PCA approach successfully reduced the feature space to two dimensions while retaining the ability to detect heterogeneity.  

\begin{figure}[h!]
    \centering
    \includegraphics[width=1\textwidth]{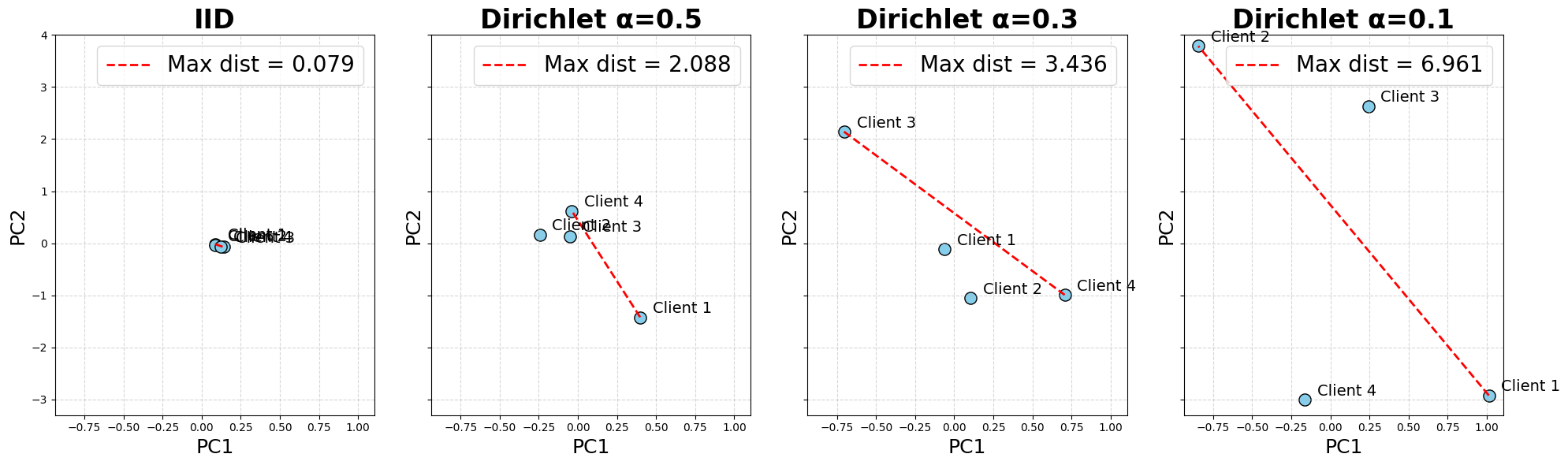}
    \caption{PCA centroid analysis on \texttt{CIFAR-10} (4 nodes) dataset.}
    \label{fig:pca_centroid_cifar10}
\end{figure}

Figure~\ref{fig:pca_centroid_cifar10} shows a similar test run on the image dataset \texttt{CIFAR-10}, with IID partitioning and Dirichlet splits at $\alpha=0.5$, $\alpha=0.3$ and $\alpha=0.1$ (lower $\alpha$ indicating greater class imbalance). Image features were extracted using ResNet-18 embeddings. Again, a clear progression is observed: the maximum pairwise distance increases from IID (0.079) to $\alpha=0.1$ (6.961). These results further demonstrate that PCA provides a robust means of reducing high-dimensional feature spaces to a low-dimensional metric, which is also capable of correctly detecting feature skew.

It could be argued that a more straightforward alternative for detecting feature skew would be to compute divergence measures (e.g., JS divergence) directly on raw feature distributions \cite{dubey2025quantifying}. However, high-dimensional spaces often contain irrelevant or noisy dimensions that obscure true distributional differences; in tabular data these may correspond to uninformative columns, and in images to pixel-level noise or background variation. Divergence computed directly in this space can therefore overestimate differences when noise dominates, or underestimate them when skew lies in a low-variance subspace. For this reason, the federated PCA mechanism was preferred, as its robustness to noise makes it more suitable for practical platforms designed to accommodate diverse datasets.

\section{Multi-Trial Optimisation}
This section presents the variant of the framework that operates under a more relaxed computational budget, optimised during \textit{Phase 3} of the development process. Unlike the single-trial mode discussed throughout the previous sections, multiple trials are permitted with this variant, enabling iterative refinement of strategy selection without approaching the full cost of hyperparameter optimisation. Two approaches were investigated before arriving to the final design of this variant. The first, \textit{multi-shot LLM prompting}, used outcomes from earlier trials to guide subsequent prompts but proved limited in efficiency. The second approach, which ultimately formed the basis of this variant, introduced a lightweight \textit{genetic search} that evolved strategies and hyperparameters through mutation and recombination. The tests that were conducted during this phase predominantly benchmarked against the HPO empirical best reference, which represents the best empirical performance after reasonable HPO trials.

\subsection{Preliminary Tests with Multi-Shot LLMs}
\begin{figure}[h!]
    \centering
    \includegraphics[width=1.0\textwidth]{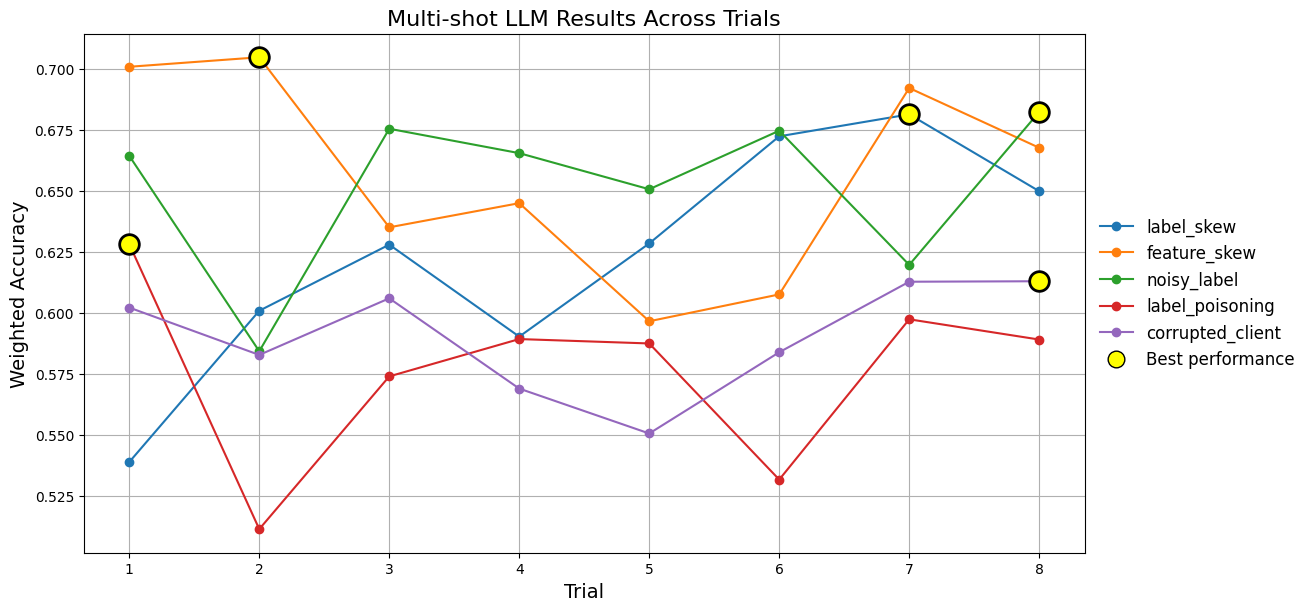} % or diagram.pdf
    \caption{LLM multi-shot performance over 8 trials on five simulated datasets.}
    \label{fig:llm_multi_random_walk}
\end{figure}

A multi-shot LLM approach was initially explored to evaluate whether an LLM could act as a hyperparameter tuner by iteratively improving its proposals based on the performance of earlier trials.  Multiple prompt templates were tested, but the most effective was one that explicitly instructed the LLM to balance exploration and exploitation. The prompt also enforced a constraint that no trial should repeat a previously attempted configuration, as duplications waste limited trial opportunities. Figure \ref{fig:llm_multi_random_walk} shows the multi-shot performance for each of the five simulated datasets over 8 trials.

The test results showed that the performance gains of the best performing trials from the first trial were minimal. Across all heterogeneity scenarios, the trial-to-trial progression resembled a random walk, with little evidence of systematic exploitation of good candidates. When visualised as a time series, the performance fluctuated around a baseline without any consistent upward trajectory, strongly suggesting that the LLM was unable to tune hyperparameters meaningfully. This led us to believe that while LLMs can identify plausible aggregation strategies in one-shot settings, they are not yet effective for hyperparameter optimisation. The extra compute required for multi-shot prompting, covering both heterogeneity detection and prompt queries, is not justified by the marginal improvements observed.

\subsection{Lightweight Genetic Search Implementation}
\label{sec:genetic-search}
A second line of investigation was to explore whether a lightweight evolutionary search method could more efficiently refine aggregation strategy parameters. Genetic algorithms were selected as they are known to balance exploration and exploitation effectively and to outperform Bayesian optimisation in scenarios with small iteration budgets.

\begin{figure}[h!]
    \centering
    \begin{minipage}{0.49\textwidth}
        \centering
        \includegraphics[width=\textwidth]{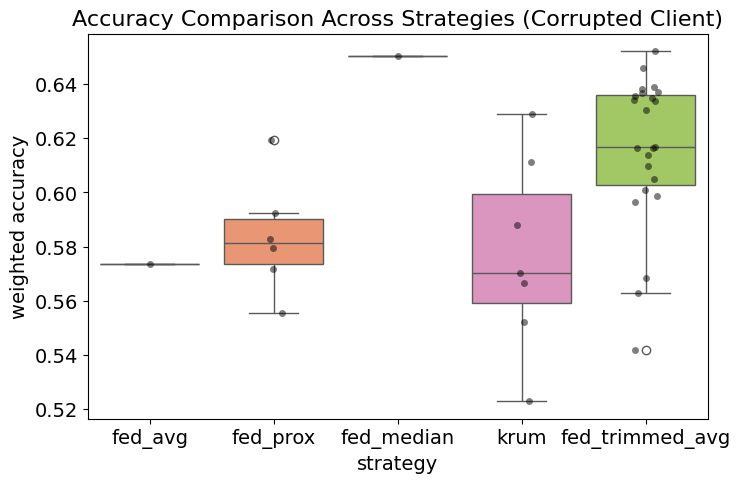}
        \caption{Strategy performance on corrupted client.}
        \label{fig:strategy_pattern_outlier}
    \end{minipage}
    \hfill
    \begin{minipage}{0.49\textwidth}
        \centering
        \includegraphics[width=\textwidth]{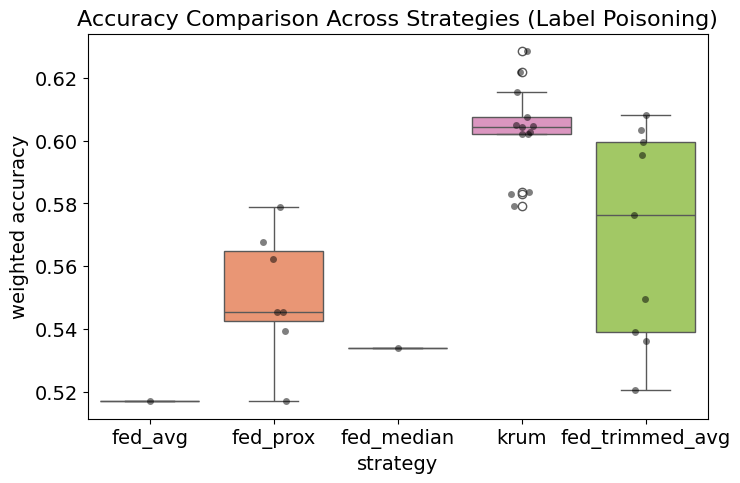}
        \caption{Strategy performance on label poisoning.}
        \label{fig:strategy_pattern_label_poisoning}
    \end{minipage}
\end{figure}

To test the feasibility of genetic algorithms, we first analysed the performance patterns from extensive hyperparameter searches conducted using Optuna, where 50 hyperparameter tuning trials were run for each heterogeneity condition. A clear pattern emerged; in most cases, a single strategy class dominated once sufficient trials had been explored. To illustrate this, Figure \ref{fig:strategy_pattern_outlier} shows that in the corrupted client heterogeneity scenario, \texttt{FedMedian} consistently outperformed alternative approaches, while \texttt{FedTrimmedAvg} was clearly superior to \texttt{Krum}. Conversely, in the label poisoning scenario (Figure \ref{fig:strategy_pattern_label_poisoning}), \textit{Krum} significantly outperformed all other strategies.

These test results informed our decision to implement the genetic search algorithm in two stages, where the first stage is to explore strategy classes to identify high-potential candidates, and the second stage is to refine parameter choices locally within the promising strategy regions. This approach allows initial populations to capture broad diversity, and successive generations to refine solutions through selection and mutation. In this way, genetic search offers a computationally efficient mechanism for balancing exploration and exploitation within the tight trial budget of federated learning. Algorithm~\ref{algo:genetic-search} below summarises the two-stage optimisation process that we implemented. 

\begin{algorithm}[h!]
\caption{Genetic search for FL aggregation strategies.}
\label{algo:genetic-search}
\begin{algorithmic}[1]
\State \textbf{Input:} generations $=2$, population size $=4$, top-$k=2$, mutation rate $=1$
\State Initialise $\text{evaluated} \gets \varnothing$, $\text{archive} \gets \varnothing$
\For{generation $g \in \{1,2\}$}
    \State Initialise candidate set $C \gets \varnothing$
    \While{$|C| < \text{population size}$}
        \If{$g = 1$}
            \State $\theta \gets \text{SampleUniform}(\Theta)$ \Comment{Random strategy and parameters}
        \Else
            \State $P \gets \text{TopK}(\text{archive}, k=2)$ \Comment{Global elites}
            \State $\text{parent} \gets \text{Uniform}(P)$
            \State $\theta \gets \text{Mutate}(\text{parent})$ \Comment{Strategy-preserving mutation}
        \EndIf
        \If{$\text{Hash}(\theta) \notin \text{evaluated}$}
            \State $\text{evaluated} \gets \text{evaluated} \cup \{\text{Hash}(\theta)\}$
            \State $C \gets C \cup \{\theta\}$
        \EndIf
    \EndWhile
    \For{each $\theta \in C$}
        \State $F(\theta) \gets \text{RunFLandScore}(\theta)$ \Comment{Weighted accuracies of last 5 FL rounds, $0$ on failure}
        \State $\text{archive} \gets \text{archive} \cup \{(\theta, F(\theta))\}$
    \EndFor
\EndFor
\State \textbf{Output:} $\theta^\star = \arg\max_{(\theta, F(\theta)) \in \text{archive}} F(\theta)$
\end{algorithmic}
\end{algorithm}

\bigskip

The search procedure aims to identify an aggregation strategy $\theta \in \Theta$ that maximises a scalar fitness function $F(\theta)$, corresponding to the downstream performance of the FL model. The algorithm begins with initialising the search space. For each candidate configuration $\theta$, the fitness $F(\theta)$ is measured as the mean of the last five weighted accuracies recorded during the FL training, which provides 
a robust proxy for converged model performance. Any configuration that fails during 
execution is assigned a fitness of zero.  The initial generation ($G=1$) is constructed by sampling four unique configurations 
uniformly at random from the search space $\Theta$. To ensure uniqueness, each 
configuration is serialised in JSON format and hashed, preventing repeated evaluation 
of identical candidates.  

Subsequent generations rely entirely on mutation. Parents are selected uniformly at random from the 
top two configurations observed across all previous evaluations, rather than being 
restricted to the most recent generation. Offspring configurations are generated 
as follows. For integer-valued parameters, mutations are 
performed by adding a perturbation $\delta \sim \text{Unif}\{-1,0,1\}$ followed by 
clamping to the parameter’s domain. For real-valued parameters, mutations take the 
form:
\\
\begin{equation}
x' = \text{clip}(x + \epsilon, [a,b]), \quad \epsilon \sim \text{Unif}[-0.1, 0.1],
\end{equation}
\\
with values rounded to four decimal places. The aggregation strategy itself remains 
fixed during mutation, with only the corresponding hyperparameters altered. As in the 
initial generation, deduplication is enforced through hashing of candidate configurations.  

The evolutionary loop proceeds for two generations, each producing four unique candidate 
configurations. All evaluated pairs $(\theta, F(\theta))$ are retained across generations. 
Upon termination, the algorithm identifies the configuration $\theta^\star$ with the maximum 
observed fitness:  
\begin{equation}
\theta^\star = \arg\max_{\theta \in \mathcal{E}} F(\theta),
\end{equation} 
where $\mathcal{E}$ represents the complete set of configurations evaluated throughout the search.  

\subsubsection{Key Design Choices of the Multi-Trial Variant}
Three principal design choices underpin the search procedure of the multi-trial variant. First, a global archive is maintained in which no configuration is repeated. Given the very limited budget of only eight trials, every evaluation represents a valuable opportunity, and strict uniqueness ensures that the search space is exploited efficiently. Secondly, parents are selected from the global archive rather than from the current generation or a restricted subset of candidates; this design encourages exploitation of the best-performing strategies identified so far, which is particularly important when the search budget is tightly constrained. Finally, the mutation rate is introduced as the probability of generating offspring through mutations of existing strategies instead of random exploration. We fixed it at $1$, such that all offspring in the second generation are produced by mutation. Lower settings such as $0.8$ or $0.5$ yielded inferior performance, reflecting that the initial generation already provides four diverse random samples, making it more effective for the second generation to focus entirely on exploiting promising strategies and exploring their surrounding parameter space.

Further tests were carried out on a hybrid design that combined LLM suggestions with the genetic search procedure. Motivated by the success of one-shot LLM prompting, which produced sensible candidate strategies, the initial generation of four random samples in the genetic search was replaced with four configurations proposed by the LLM. The results indicated that the second generation of candidates successfully concentrated on highly promising regions of the search space, consistent with the expected behaviour of genetic search. 

However, the best-performing configurations achieved in this hybrid setting were only comparable to those identified by the pure genetic search, and in no instance did the hybrid approach yield a configuration that significantly outperformed the genetic search approach. Furthermore, the hybrid approach introduced additional costs, both computational and financial. In particular, it incorporated the heterogeneity detection process and incurred expenses associated with LLM API usage. As no improvement in performance was observed, these additional costs could not be justified. Consequently, the approach selected for multi-trial settings was the genetic search implementation excluding the LLM.

\subsection{Key Takeaways}
The findings of \textit{Phase 3} suggest that LLMs, while effective in one-shot reasoning over strategy families, are not yet reliable tools for hyperparameter optimisation in compute-constrained FL strategy selection. Within a limited trial budget, multi-trial prompting produced outcomes resembling random search rather than systematic tuning. Our preferred approach was to therefore make use of a carefully designed genetic search method as an alternative. Our approach achieved performance close to the HPO empirical best while requiring only 8 trials, demonstrating that FL strategy selection can be optimised efficiently within realistic compute constraints. Whilst the hybrid approach that combined LLMs and genetic search was disregarded due to inefficiency, it yielded an important insight; genetic search can refine the search space initially suggested by LLMs, thereby compensating for their weakness in hyperparameter tuning while leveraging their strength in proposing plausible configurations for further refinement.

\section{Evaluation}
\label{chp9:eval}
The previous sections presented the features of the framework and detailed how each component was optimised. Here, in \textit{Phase 4}, we shift the focus to evaluating the framework as a whole, assessing its scalability and generalisability. We begin by evaluating the framework on the development setups. The scope is then progressively expanded to examine whether the performance from development remains consistent under more demanding conditions: namely increasing numbers of participating client nodes, and varying data modalities spanning tabular, image, text, and reinforcement learning tasks.

\subsection{Evaluation on the Development Dataset}
\label{sec:devset-eval}
We first evaluate the framework on the development datasets, assessing downstream FL predictive performance and runtime for both the single-trial LLM recommendation and the genetic search approach.

\subsubsection{Downstream FL Performance: Automated Heterogeneity Detection}
To assess whether automated heterogeneity detection improves FL performance, we conducted an evaluation by comparing the LLM-recommended aggregation strategy against FedAvg. For each heterogeneity scenario, both methods were run ten times, and mean test accuracy and standard deviation were reported to account for stochasticity.

\begin{table}[h!]
\centering
\small
\begin{tabular}{lcc}
\toprule
\textbf{Simulated Heterogeneity} & \textbf{LLM Recommended Strategy} & \textbf{FedAvg} \\
\midrule
% xxx & mean $\pm$ std & mean $\pm$ std \\
Label Skew & \textbf{0.6360} $\pm$ 0.0058 & 0.6186 $\pm$ 0.0168 \\
Label Poisoning & \textbf{0.6087} $\pm$ 0.0180 & 0.5543 $\pm$ 0.0258 \\
Noisy Label & \textbf{0.6548} $\pm$ 0.0410 & 0.6393 $\pm$ 0.0500 \\
Feature Skew & \textbf{0.7022} $\pm$ 0.0241 & 0.6806 $\pm$ 0.0347 \\
Corrupted Client & \textbf{0.6265} $\pm$ 0.0219 & 0.5948 $\pm$ 0.0197 \\
IID & 0.7208 $\pm$ 0.0200 & 0.7253 $\pm$ 0.0261 \\
\midrule
CIFAR10 (Dirichlet $\alpha=0.1$) & \textbf{0.2604} $\pm$ 0.0115 & 0.2121 $\pm$ 0.0266 \\
CIFAR10 (Dirichlet $\alpha=0.3$) & \textbf{0.3723} $\pm$ 0.0278 & 0.3508 $\pm$ 0.0153 \\
\bottomrule
\end{tabular}
\caption{Downstream FL weighted accuracy (mean $\pm$ std over 10 runs) comparing LLM recommendation and FedAvg.}
\label{tab:oneshot_vs_fedavg}
\end{table}

Table~\ref{tab:oneshot_vs_fedavg} shows that the recommended strategy consistently outperforms FedAvg under heterogeneous conditions. The largest improvement occurs in the label poisoning scenario, indicating that the framework effectively detects adversarial clients and selects robust aggregation strategies. Gains are also observed for label skew, noisy labels, feature skew, and corrupted clients, demonstrating broad applicability across heterogeneity types. In the IID setting, both approaches achieve comparable performance, indicating that the automated workflow does not degrade performance when heterogeneity is absent.

Across all runs, the LLM generated valid aggregation configurations without syntax or integration errors, confirming reliable end-to-end deployment within the FL pipeline. These results support the hypothesis that LLMs can recommend qualitatively appropriate FL strategies in a single trial.

Experiments on CIFAR-10 further validate the approach beyond tabular data. Under Dirichlet partitioning with $\alpha = 0.1$ and $\alpha = 0.3$, the recommended strategies yielded consistent accuracy improvements over FedAvg, indicating that the heterogeneity detection and strategy selection pipeline remains effective for high-dimensional image representations.

Although the absolute gains are modest, this is expected given the simplicity of the binary NASA Bearing dataset and lightweight models, which limit the potential benefit of aggregation strategy adaptation. Nonetheless, the consistent improvements across heterogeneous scenarios demonstrate that automated heterogeneity detection can meaningfully enhance FL robustness. More complex datasets and models are evaluated in Section~\ref{chp9_scale_nodes}.

\subsubsection{Downstream FL Performance: Lightweight Genetic Search}
A similar evaluation was conducted for the lightweight genetic search module. In addition to FedAvg and the single-shot LLM recommendation, we also benchmark an empirical best and empirical worst reference obtained via Optuna-based hyperparameter optimisation with 50 trials, representing upper and lower performance bounds under relaxed computational budgets.

\begin{table}[h!]
\centering
\small
\begin{tabular}{lccccc}
\toprule
\textbf{Simulated Heterogeneity} & \textbf{Empirical Best} & \textbf{Genetic Search} & \textbf{Single-Shot LLM} & \textbf{FedAvg} & \textbf{Empirical Worst}\\
\midrule
Label Skew  & 0.6729 $\pm$ 0.0149 & 0.6679 $\pm$ 0.0128 & 0.6360 $\pm$ 0.0058 & 0.6186 $\pm$ 0.0168  & 0.6236 $\pm$ 0.0205 \\
Label Poisoning & 0.6060 $\pm$ 0.0168 & 0.6223 $\pm$ 0.0195 & 0.6087 $\pm$ 0.0180 & 0.5543 $\pm$ 0.0258 & 0.5543 $\pm$ 0.0258 \\
Noisy Label & 0.6948 $\pm$ 0.0187 & 0.6882 $\pm$ 0.0233 & 0.6548 $\pm$ 0.0410 & 0.6393 $\pm$ 0.0500 & 0.6434 $\pm$ 0.0471 \\
Feature Skew  & 0.7250 $\pm$ 0.0139 & 0.7112 $\pm$ 0.0449 & 0.7022 $\pm$ 0.0241 & 0.6806 $\pm$ 0.0347 & 0.6681 $\pm$ 0.0429 \\
Corrupted Client & 0.6058 $\pm$ 0.0258 & 0.6120 $\pm$ 0.0198 & 0.6265 $\pm$ 0.0219 & 0.5948 $\pm$ 0.0197 & 0.5838 $\pm$ 0.0230 \\
\bottomrule
\end{tabular}
\caption{Downstream FL weighted accuracy (mean $\pm$ std over 10 runs) comparing genetic search, LLM recommendation, FedAvg, and empirical HPO bounds.}
\label{tab:ch8_genetic_performance}
\end{table}

Table~\ref{tab:ch8_genetic_performance} summarises the downstream FL performance on the development dataset. The genetic search consistently approaches the empirical best reference and surpasses it in some scenarios. In most cases, genetic search also outperforms the single-shot LLM recommendation, suggesting that precise hyperparameter tuning may remain a limitation for one-shot prompting. By contrast, the genetic search effectively refines hyperparameters and achieves performance close to the empirical optimum under significantly reduced computational cost.

\subsubsection{Runtime and Complexity Analysis}
\paragraph{Automated Heterogeneity Detection}
Having demonstrated the utility of heterogenenity detection in single-trial settings, we now provide a reference summary of the runtime costs of the full heterogeneity detection pipeline, expressed in terms of equivalent rounds of naïve FL aggregation strategy trials. Table~\ref{tab:fededa_runtime} reports the runtime of each detection component on the \texttt{NASA Bearings} dataset. The entire heterogeneity detection procedure completes in 73.97 seconds. For reference, this corresponds to the cost of approximately 63 standard FL rounds, each of which requires about 1.16 seconds. On this dataset, the FL model typically converges within roughly 30 rounds, so the additional overhead amounts to only about two extra naïve training runs. This is an attractive trade-off for obtaining a completely automated and informed strategy selection upfront. In contrast, without our framework, tuning strategy hyperparameters would require multiple full FL executions.

\begin{table}[h!]
\centering
\small
\begin{tabular}{l r}
\hline
\textbf{Component} & \textbf{Time (s)} \\
\hline
\textit{Reference: One FL Round}   & \textit{1.1613} \\
Label skew detection module     & 8.3163 \\
Feature skew detection module  & 11.1986 \\
Outlier detection module     & 54.4542 \\
Total Time: Full heterogeneity detection process            & 73.9691 \\
\hline
\end{tabular}
\caption{Runtime of heterogeneity detection components.}
\label{tab:fededa_runtime}
\end{table}

In terms of component breakdown, the outlier detection module dominates the computational cost (54.5s), since it directly involves FL training and scales with the training workload. However, this cost is still substantially lower than running multiple naïve trials, and early identification of adversarial behaviour can prevent downstream degradation. 

Feature skew detection module is the second-largest contributor (11.2s), with cost scaling primarily with the feature dimension $d$ and the number of clients $n$ (complexity analysis performed below). In this case, the federated PCA step is equivalent to just 9.6 FL rounds, substantially less than a single naïve trial. Moreover, as feature spaces grow more complex, FL convergence typically requires more rounds, making the relative cost of feature skew detection increasingly negligible. That said, the method still scales with $d$, and since $n$ is an uncontrolled variable, reducing feature dimensionality during preprocessing is advisable where possible. For instance, when extracting features from image data, it is preferable to use lower-dimensional embeddings to improve efficiency.

In terms of PCA complexity; on a client $i$ with $n_i$ samples and $d$ features, local sufficient statistics in the first round require $\mathcal{O}(n_i d)$ operations for per-feature sums and $\mathcal{O}(n_i d^2)$ for cross-product terms whilst constructing the covariance, the latter dominating the workload. In the second round, projecting onto two principal components costs $\mathcal{O}(n_i d)$, which is comparatively minor. Summed across clients, the overall client-side complexity is $\mathcal{O}(N d^2)$ with $N=\sum_i n_i$. On the server side, aggregation requires combining the $\mathcal{O}(d^2)$ covariance contributions received from each of the $K$ clients, giving a total of $\mathcal{O}(K d^2)$. Once the full $d \times d$ covariance matrix is assembled, PCA is performed. A full eigendecomposition of this matrix has complexity $\mathcal{O}(d^3)$, while iterative methods that only compute the top-$k$ components (with $k=2$ in this study) can reduce this to approximately $\mathcal{O}(d^2)$. Thus, the server cost scales with the number of clients $K$ and quadratically with feature dimension $d$, but is independent of the total number of samples $N$.  

Overall, the runtime cost grows linearly with $N$ (on the client side) and quadratically with $d$ (on both client and server sides). The dominant term is typically the client-side $\mathcal{O}(n_i d^2)$ cross-product computation, while server costs become significant only when $d$ is very large; reducing $d$ during feature selection or embedding is therefore the most effective way to improve efficiency.

\paragraph{Comparative Study of Each Approach }
Runtime was also recorded for each proposed approach alongside reference baselines of a single FL training trial and Optuna with 50 trials, as reported in Table~\ref{tab:approach_eval_runtime}. Each trial is run for 30 rounds, corresponding to the point of convergence on this dataset. The runtime analysis suggests that improved performance observed in Table~\ref{tab:ch8_genetic_performance} does not come at a prohibitive cost; single-trial LLM automation adds only the equivalent of about one extra FL training round, while genetic search remains far cheaper than Optuna due to the capped number of trials. This balance of predictive gains and modest runtime overhead makes the freamework highly practical for strategy selection under various compute budgets.

\begin{table}[h!]
\centering
\small
\begin{tabular}{l r}
\hline
\textbf{Approach} & \textbf{Time (s)} \\
\hline
\textit{Reference: One Naive FL Training (30 FL Rounds)}  & \textit{69.211} \\
One-Trial LLM Automation & 109.256 \\
Genetic Search &  251.805\\
Optuna (50 Trials, 30 Rounds Each) & 1105.918 \\
\hline
\end{tabular}
\caption{Runtime analysis of different FL strategy selection approaches.}
\label{tab:approach_eval_runtime}
\end{table}

\subsection{Scaling to Larger Numbers of Nodes}
\label{chp9_scale_nodes}
We then evaluated the scalability of the proposed framework as the number of participating nodes increased under varying degrees of data skew. Using \texttt{CIFAR-10}, the number of nodes is scaled (4, 10, 50, 100) under Dirichlet partitions~\cite{yurochkin2019bayesian} with $\alpha \in {0.1,0.3,0.5}$, while model architecture and training parameters are kept fixed. Smaller $\alpha$ values create more imbalanced label distributions across clients, which in image data also induces feature skew due to differing category distributions. Both performance and runtime were assessed for each setting.

\begin{figure}[!htbp]
\centering
  \includegraphics[width=1.0\textwidth]{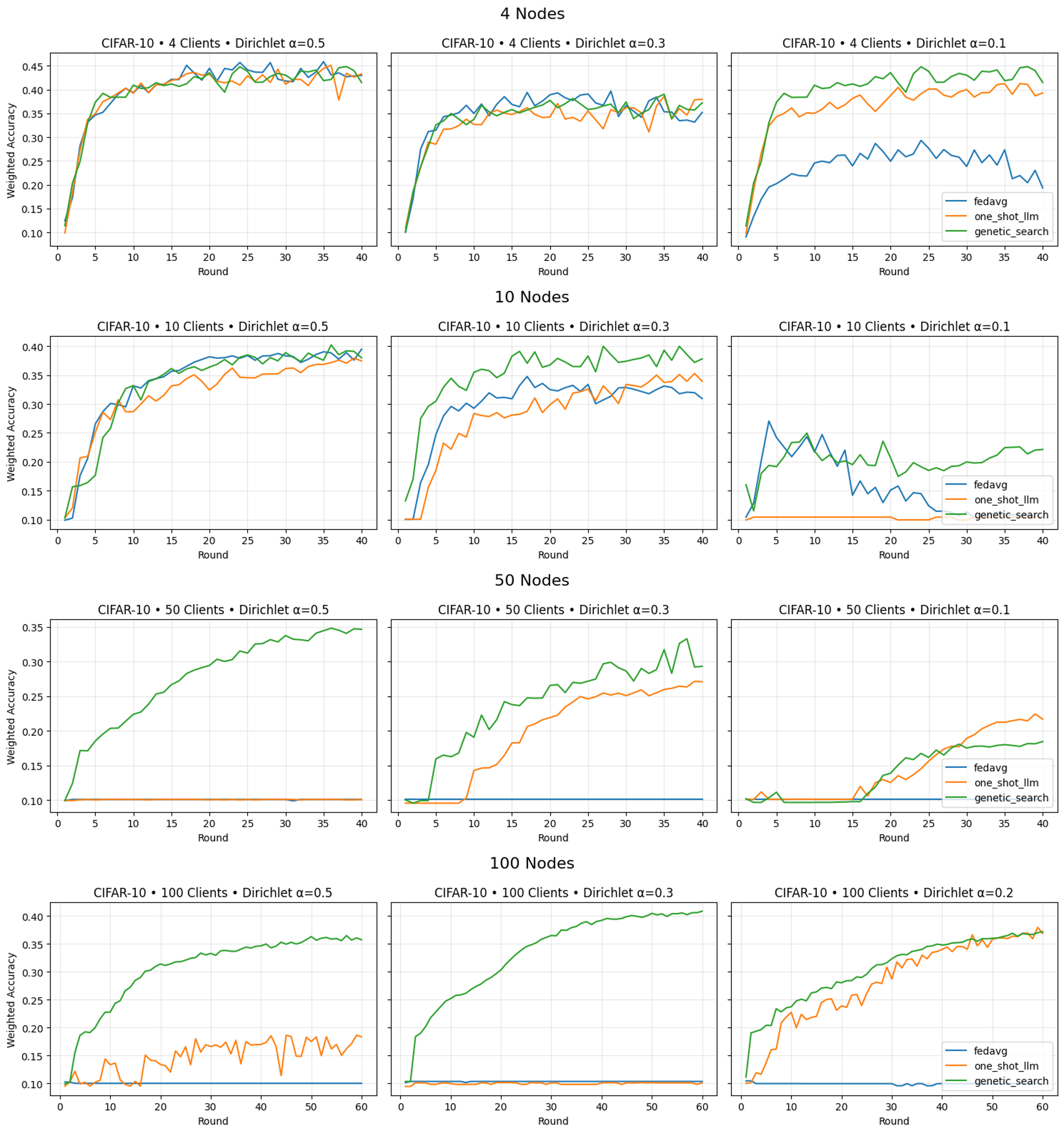}
  \caption{Performance analysis on increasing number of nodes and varying Dirichlet partitioning on \texttt{CIFAR-10.}}
  \label{fig:cifar-scaling}
\end{figure}

Figure~\ref{fig:cifar-scaling} summarises the performance results, where consistent patterns emerged. Genetic search (green) always achieved the highest performance, the single-trial LLM selector (orange) provided modest gains over \textit{FedAvg} in some cases, and \textit{FedAvg} (blue) underperformed in most cases. This ordering holds across all node counts, demonstrating the robustness of the proposed approach under scaling. In low-node settings (4 and 10 nodes), differences were small when $\alpha=0.5$ (close to IID), with \textit{FedAvg} occasionally matching or slightly exceeding the one-trial LLM. However, at higher skew (Dirichlet $\alpha=0.1$), gaps quickly emerged, with both automated approaches outperforming \textit{FedAvg}.  

As the number of nodes grows, the benefits of genetic search become more pronounced. At 50 and 100 nodes, both \textit{FedAvg} and the one-trial LLM frequently stagnate. \textit{FedAvg} remains flat in all settings, while the one-trial LLM shows occasional stagnation, particularly at 50 clients with $\alpha=0.5$ and 100 clients with $\alpha=0.3$. This behaviour reflects the growing difficulty of collaborative learning as the data is partitioned across many nodes and class imbalance becomes more severe. In contrast, genetic search identifies the few viable strategies that continue to train successfully, even when most candidate strategies fail. This was particularly clear at 100 nodes, where only genetic search shows upward trajectories while \textit{FedAvg} remains flat and the LLM achieves limited progress, further reinforcing the effect of genetic search. Overall, these results confirm that the framework scales effectively with the number of nodes. Genetic search consistently maximises performance under challenging settings, while the one-shot LLM provides lightweight yet noticeable improvements at minimal computational cost.  

\begin{figure}[H]
\centering
  \includegraphics[width=0.9\textwidth]{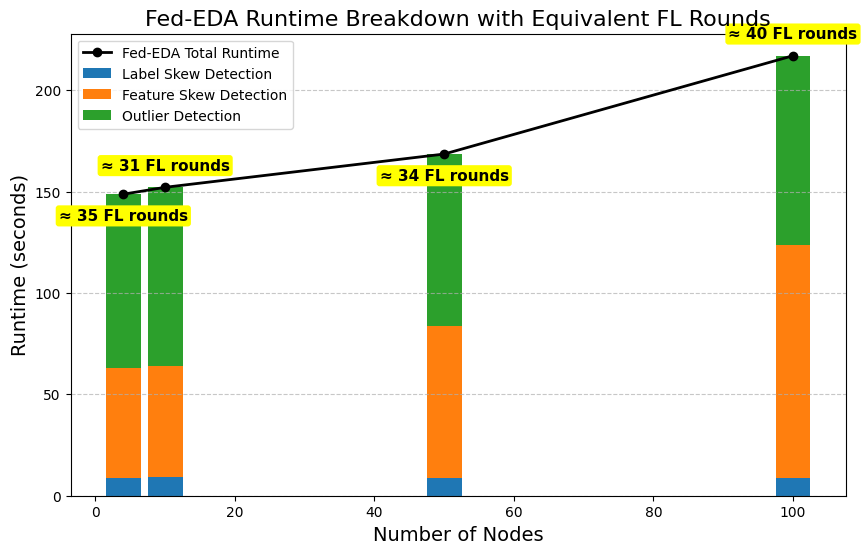}
  \caption{Heterogeneity detection runtime analysis on increasing number of nodes.}
  \label{fig:runtime_anlaysis}
\end{figure}

We also carried out a runtime analysis on the heterogeneity detection module, as non-trivial overheads come from heterogeneity diagnosis, which adds to the one-trial LLM approach on top of a single FL training round. Figure~\ref{fig:runtime_anlaysis} shows how heterogeneity detection runtime scales with the number of nodes. On the same hardware, total runtime increases from 148.7s (4 nodes) to 152.1s (10 nodes), 168.5s (50 nodes), and 217.0s (100 nodes). The corresponding per-round costs are 4.859s, 4.867s, 5.300s, and 51.233s. In terms of equivalent FL rounds, the detection process corresponds to 35, 31, 34, and 40 rounds, respectively. For this model, convergence requires about 40 rounds for 4–50 nodes and 60 rounds for 100 nodes. Thus, the overhead of heterogeneity detection is roughly equivalent to a single additional round of naïve parameter tuning, a modest cost for giving the LLM sufficient context for informed one-shot selection. As node count increases, FL convergence takes longer, while the relative overhead of detection diminishes, making the process increasingly justifiable at larger scales.

Breaking down the detection process by component shows that label skew detection remains almost constant as the number of nodes increases, whereas feature skew and outlier detection scale with node count. Feature skew detection grows with node count, while outlier detection scales linearly with per-round FL training and therefore with the training workload. Future refinements could focus on improving the efficiency of these components. Nonetheless, in the current setup, overall runtime remains within a justifiable range compared to the benefits provided and the alternative compute-intensive fine-tuning process.

\subsection{Generalisation Across Data Modalities}
\label{chp9_other_data_types}
\begin{figure}[H]
\centering
  \includegraphics[width=0.9\textwidth]{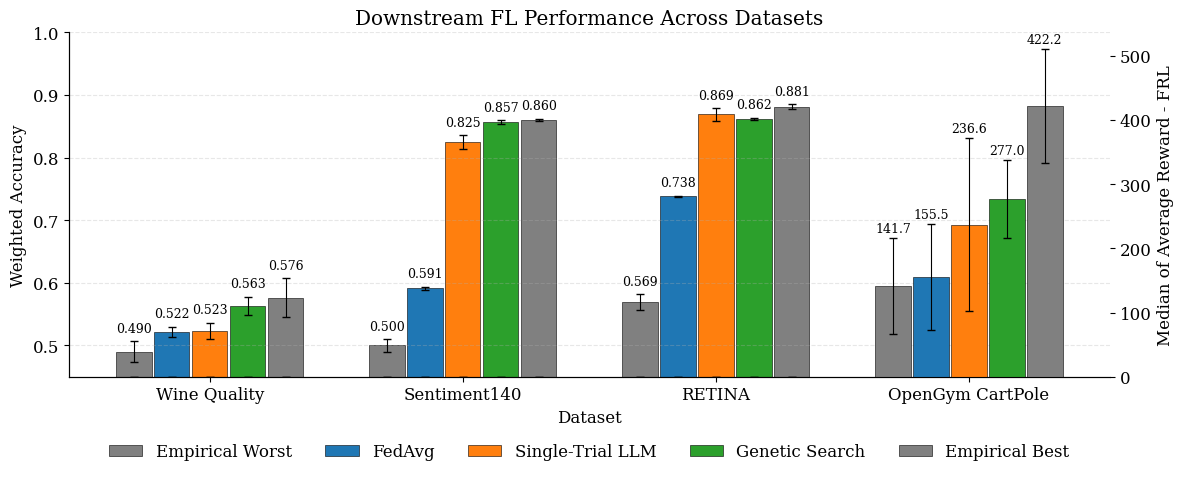}
  \caption{Performance comparison of our framework against several baselines across different datasets. Weighted accuracy is reported for all datasets, except for OpenGym \texttt{CartPole} which uses median average reward.}
  \label{fig:scale_data_anlaysis}
\end{figure}

The framework was also tested across diverse data modalities, including tabular datasets with multiple labels, image and text data, and finally a federated reinforcement learning (FRL) task on the OpenGym \texttt{CartPole} Environment. Each dataset demanded a distinct ML architecture, providing a natural test of the framework’s flexibility. Figure~\ref{fig:scale_data_anlaysis} summarises performance across all datasets. The same pattern that we observed during development held, where the one-trial LLM approach consistently outperformed \textit{FedAvg} by a small amount, while genetic search achieved the best results and closely matched the HPO empirical best reference. This led us to believe that the proposed framework scales effectively across different modalities. Detailed insights from each dataset are summarised below. Note that for FRL, performance was measured as the median of the average reward across devices, obtained by repeating each strategy configuration five time, as opposed to measuring the weighted average accuracy. This adjustment was done to account for the high stochasticity of RL, thereby reducing sensitivity to outlier trials.

With the \texttt{Wine Quality Dataset}, we extended the tabular data evaluation to a multi-class task. Improvements were modest, but the same improvement trends of the two approaches confirmed generalisability beyond binary classification. In the case of the \texttt{Retina Dataset}, heterogeneity arose from real-world factors such as varying hospital imaging equipment and patient demographics; both automated approaches substantially improved over \textit{FedAvg}, with the one-trial LLM achieving results close to the HPO empicial best and genetic search adding marginal gains, highlighting the framework’s robustness in real-world, non-simulated heterogeneity scenarios. We then extended the evaluation by using a TextCNN model and testing it on the \texttt{Sentiment140 Dataset}. The consistent performance pattern indicated robustness to high-dimensional text data and alternative model architectures. However, results also revealed an important limitation: performance remained highly dependent on the choice of local ML model, which lies outside the scope of this project. When stress-testing with the same dataset but using a transformer setup with TinyBERT embeddings, all Optuna trials produced identical training curves, even after 300 rounds.  Investigation suggested that the pre-trained embeddings already encode rich semantic knowledge, making the feature difference between the partitioned \texttt{Sentiment140} Dataset negligible relative to the strong representations learned from large corpora of text. As a result, FL aggregation strategies showed minimal variation, since no mitigation can address the lack of discriminative features extracted by the embeddings. These findings emphasise that while the framework adapts strategies effectively under heterogeneity, it cannot compensate for unsuitable model design, pointing to a direction for future research to complement our proposed framework.

Finally, the framework was evaluated for federated RL tasks using the OpenGym\texttt{ CartPole} environment and a DQN model~\cite{mnih2013playing}. The task is to balance a pole for up to 500 timesteps, measured as rewards, by applying directional force to the cart under physics-based dynamics, parameterised as environment variables. Most devices used the default settings, while one device was configured with extreme gravity (20 m/s\(^2\)) and a shortened pole length (0.1 not 1.0), producing markedly different learning trajectories. As expected, both automated approaches outperformed \textit{FedAvg}, confirming adaptability to RL tasks. Performance is reported as the median of the average reward across five repeated runs to account for stochasticity, with the maximum reward of 500 indicating successful balancing on all devices. This extension demonstrates feasibility beyond classification and provides insight into how the pipeline can be adapted for varying tasks. While further refinement and testing are needed, these results highlight the potential of extending the framework to a broader set of ML applications with minimal engineering effort.

\subsection{Key Takeaways}
This section demonstrated that the framework reliably improves upon \textit{FedAvg} across a range of settings, including varying number of nodes and data modalities. The single-trial LLM approach offers measurable gains with minimal overhead, while the genetic search achieves performance close to, and occasionally exceeding, the HPO empirical best reference at a fraction of the cost. Runtime analyses confirmed that the additional compute required for heterogeneity detection remains modest and increasingly justifiable as system scale grows. Moreover, tests across multiple data modalities and client scales highlight the tool’s robustness and adaptability. These findings establish the generalisability of the proposed solution as a step towards more automated FL deployments.

\section{Summary and Future Work}

This paper proposed an automated approach for selecting FL aggregation strategies that are well-suited to the statistical heterogeneity of edge data, while reducing reliance on deep FL expertise and avoiding costly trial-and-error experimentation. This was realised through the development of an end-to-end framework that supports strategy selection under two compute-constrained scenarios, with flexibility for both human-in-the-loop and fully automated configurations. For single-trial scenarios, LLMs were leveraged to map either human-provided descriptions of statistical heterogeneity or automatically detected heterogeneity signals into valid and executable strategy configurations. For multi-trial scenarios, a lightweight, carefully designed genetic search method was implemented, offering an efficient optimisation mechanism that approaches the performance of exhaustive search methods such as Optuna, but with a fraction of the computational cost. In addition, novel heterogeneity detection modules were developed to diagnose label skew, feature skew, and outlier clients. These modules were lightweight and privacy-preserving. Together, this establishes a unified framework for automated aggregation strategy selection, which has not been addressed in prior work, and adapts to different levels of user involvement and compute constraints.

This work demonstrates a robust automation integrated into real-world FL systems for aggregation strategy selection. It highlights the effectiveness, strengths and limitations of each proposed approach, and provides complementary techniques to address gaps in prior work. Nevertheless, suboptimal model architectures can offset the gains from improved strategy selection. Extending automation to model selection and configuration is therefore a natural next step, paving the way toward a fully automated FL pipeline.

Beyond strategy and model-level automation, the framework exhibits strong robustness and generalisability, as evidenced by scalability studies across different data modalities. Building on this foundation, future work should extend support to a broader range of machine learning tasks and more diverse data types, ensuring reliable deployment in varied real-world settings. A current limitation is that the feature skew detection module remains sensitive to preprocessing steps, such as compression, which can distort divergence estimates. Future work could refine this mechanism to better handle such artefacts and include broader validation across datasets to guarantee stability across tasks with differing feature characteristics.

By default, FL preserves data locality and prevents the transfer of raw client data, thereby aligning with the privacy-preserving principles. Our framework strengthens this foundation by automating aggregation strategy selection, improving FL performance under non-IID conditions. Since data heterogeneity is a common barrier to effective deployment, enhancing robustness in this way directly supports wider adoption of privacy-preserving machine learning, reducing the need for centralised data sharing in sensitive domains. In addition to advancing privacy-preserving machine learning, the proposed approaches were designed with efficiency in mind. The compute-constrained approaches, particularly the one-shot and lightweight multi-trial optimisation pipelines, offer an alternative to conventional hyperparameter tuning, which is often prohibitively expensive in federated settings. 
\\\\\\

\newpage
\section{Appendix}
\label{Appendix}

This Appendix provides more information on how non-IID conditions were created.

\textit{1) Manual simulation:} 

Manual simulation provides explicit control over the type and severity of heterogeneity by directly altering data distributions. This approach is particularly valuable for isolating specific heterogeneity types (e.g., label skew vs. feature skew) and creating adversarial scenarios for testing robustness. We implemented this simulation on the \texttt{NASA Bearing} Dataset and the \texttt{CartPole} Environment as detailed below. 

For the \texttt{NASA Bearing} dataset, five varying heterogeneity scenarios were simulated. {\textit{a) Heterogeneity-free IID setting:}} In this simulation, the data points were randomly distributed and assigned so that each client reflected the global distribution; {\textit{b) Label Skew:}} Here, we simulated label skew by distributing the data across four clients with imbalanced label proportions, and each client held 90\%, 70\%, 50\%, or 10\% of a single label class;  {\textit{c) Feature Skew:} } We then simulated feature skew by evenly splitting the data among four clients and adding Gaussian noise to create distributional shifts. Noise parameters (mean, standard deviation) were set as (0, 0.1), (0, 0.5), (1, 0.1), (-0.1, 0.1); {\textit{d) Noisy Labels via Random Flipping:}} For one client, 30\% of labels were randomly flipped to simulate annotation errors, as per \cite{Li2021NonIID}; {\textit{e) Adversarial Client and Label Poisoning:}} One client was adversarial, with all labels flipped to simulate a targeted poisoning attack; {\textit{f) Corrupted Client:}} This was a combined scenario where one client had Gaussian noise added to features (mean = 1, std = 0.5) alongside complete label flipping, producing a severely corrupted outlier. For the \texttt{CartPole} Environment, heterogeneity was introduced by modifying environment parameters on one client. While most clients operated under default conditions (gravity$=9.8\,m/s^2$, pole length = 1), one client was configured with gravity $=20\,m/s^2$ and pole length = 0.1. This created distinct learning dynamics and heterogeneous model updates.

\textit{2) Partitioning-based Simulation:}

Partitioning is a widely used technique in FL to create heterogeneous data distributions without altering the raw samples. By splitting datasets according to statistical rules, clients receive non-IID subsets that mimic realistic imbalances. We employed Dirichlet partitioning \cite{yurochkin2019bayesian}, where the concentration parameter $\alpha$ controls the severity of imbalance: smaller $\alpha$ values produce more skewed class distributions, while larger values approximate IID. For \texttt{CIFAR-10}, \texttt{Wine Quality}, and \texttt{Sentiment140}, these datasets are partitioned using different $\alpha$ values to simulate varying degrees of label imbalance and distribution skew. This method enables experiments that test scalability and robustness under controlled levels of heterogeneity severity.

\textit{3) Real-World Heterogeneity:}

Unlike simulations, real-world heterogeneity arises naturally due to differences in how data are collected across client devices; for the \texttt{Retinal Images} dataset, images originated from hospitals in different regions, and variation arose from differences in imaging devices, patient populations, and clinical protocols. In our FL setup, each client device was assigned data from a specific clinic or hospital, directly reflecting the distributional differences seen in practice.

\newpage
%% bibliography
\bibliographystyle{vancouver}
\bibliography{references}

@inproceedings{mcmahan2017communication,
  title={Communication-efficient learning of deep networks from decentralized data},
  author={McMahan, H Brendan and Moore, Eider and Ramage, Daniel and Hampson, Seth and Arcas, Blaise Aguera y},
  booktitle={Artificial Intelligence and Statistics},
  pages={1273--1282},
  year={2017},
  eprint={1602.05629},
  archiveprefix={arXiv},
  primaryclass={cs.LG}
}

@article{Ye2024HeterogeneousFL,
  author  = {Ye, Mang and Fang, Xiuwen and Du, Bo and Yuen, Pong C. and Tao, Dacheng},
  title   = {Heterogeneous Federated Learning: State-of-the-Art and Research Challenges},
  journal = {ACM Computing Surveys},
  volume  = {56},
  number  = {3},
  pages   = {1--44},
  year    = {2024},
  month   = {March}
}

@article{Zhao2018NonIID,
  author  = {Zhao, Yue and Li, Meng and Lai, Liangzhen and Suda, Naveen and Civin, Jason and Chandra, Vikas},
  title   = {Federated Learning with Non-IID Data},
  journal = {arXiv preprint arXiv:1806.00582},
  year    = {2018},
  url     = {https://arxiv.org/abs/1806.00582}
}

@inproceedings{karimireddy2020scaffold,
  title={SCAFFOLD: Stochastic Controlled Averaging for Federated Learning},
  author={Karimireddy, Sai Praneeth and Kale, Satyen and Mohri, Mehryar and Reddi, Sashank and Stich, Sebastian U and Suresh, Ananda Theertha},
  booktitle={Proceedings of the 37th International Conference on Machine Learning (ICML)},
  pages={5132--5143},
  year={2020}
}

@misc{gao2022surveyheterogeneousfederatedlearning,
      title={A Survey on Heterogeneous Federated Learning}, 
      author={Dashan Gao and Xin Yao and Qiang Yang},
      year={2022},
      eprint={2210.04505},
      archivePrefix={arXiv},
      primaryClass={cs.LG},
      url={https://arxiv.org/abs/2210.04505}, 
}

@misc{beutel2022flower,
      title={Flower: A Friendly Federated Learning Research Framework}, 
      author={Daniel J. Beutel and Taner Topal and Akhil Mathur and Xinchi Qiu and Javier Fernandez-Marques and Yan Gao and Lorenzo Sani and Kwing Hei Li and Titouan Parcollet and Pedro Porto Buarque de Gusmão and Nicholas D. Lane},
      year={2022},
      eprint={2007.14390},
      archivePrefix={arXiv},
      primaryClass={cs.LG},
      url={https://arxiv.org/abs/2007.14390}, 
}

@misc{shen2023llmedgeai,
      title={Large Language Models Empowered Autonomous Edge AI for Connected Intelligence}, 
      author={Yifei Shen and Jiawei Shao and Xinjie Zhang and Zehong Lin and Hao Pan and Dongsheng Li and Jun Zhang and Khaled B. Letaief},
      year={2023},
      eprint={2307.02779},
      archivePrefix={arXiv},
      primaryClass={cs.IT},
      url={https://arxiv.org/abs/2307.02779}, 
}

@misc{citil2023nasa,
  author       = {Furkan Çitil},
  title        = {NASA Bearing Dataset - Supervised Learning},
  year         = {2023},
  howpublished = {\url{https://www.kaggle.com/code/furkancitil/nasa-bearing-dataset-supervised-learning}},
  note         = {Accessed: 2025-05-31}
}

@inproceedings{dubey2025quantifying,
  title={Quantifying and Analyzing Client Data Heterogeneity in Federated Learning via Multi-Modal Divergence Metrics},
  author={Dubey, Praveer and Kumar, Mohit},
  booktitle={TechRxiv Preprint},
  year={2025}
}

@article{kummaya2025fedhetero,
  title={Fed-Hetero: A Self-Evaluating Federated Learning Framework for Data Heterogeneity},
  author={Milan Kummaya, Aiswariya and Joseph, Amudha and Rajamani, Kumar and Ghinea, George},
  journal={Applied System Innovation},
  volume={8},
  number={2},
  pages={28},
  year={2025},
  publisher={MDPI},
  doi={10.3390/asi8020028}
}

@article{Li2021NonIID,
  author    = {Li, Qinbin and Diao, Yiqun and Chen, Quan and He, Bingsheng},
  title     = {Federated Learning on Non-IID Data Silos: An Experimental Study},
  journal   = {arXiv preprint arXiv:2102.02079},
  year      = {2021},
  url       = {https://doi.org/10.48550/arXiv.2102.02079}
}

@article{Qi2023ModelAggregationSurvey,
  author    = {Qi, Pian and Chiaro, Diletta and Guzzo, Antonella and Ianni, Michele and Fortino, Giancarlo and Piccialli, Francesco},
  title     = {Model Aggregation Techniques in Federated Learning: A Comprehensive Survey},
  journal   = {Future Generation Computer Systems},
  year      = {2024},
  volume    = {150},
  pages     = {272--293}
}

@ARTICLE{mawela2025webbasedFL,
  author={Mawela, Chamith and Issaid, Chaouki Ben and Bennis, Mehdi},
  journal={IEEE Internet of Things Journal}, 
  title={A Web-Based Solution for Federated Learning With LLM-Based Automation}, 
  year={2025},
  volume={12},
  number={12},
  pages={19488-19503},
  doi={10.1109/JIOT.2025.3542897}}

@inproceedings{akiba2019optuna,
  title     = {Optuna: A Next-Generation Hyperparameter Optimization Framework},
  author    = {Akiba, Takuya and Sano, Shotaro and Yanase, Toshihiko and Ohta, Takeru and Koyama, Masanori},
  booktitle = {The 25th ACM SIGKDD International Conference on Knowledge Discovery \& Data Mining},
  pages     = {2623--2631},
  year      = {2019},
}

@techreport{krizhevsky2009learning,
  title       = {Learning multiple layers of features from tiny images},
  author      = {Krizhevsky, Alex and Hinton, Geoffrey},
  year        = {2009},
  institution = {University of Toronto},
  type        = {Technical Report},
  number      = {TR-2009},
  publisher   = {Citeseer}
}

@inproceedings{yurochkin2019bayesian,
  title     = {Bayesian Nonparametric Federated Learning of Neural Networks},
  author    = {Yurochkin, Mikhail and Agarwal, Mayank and Ghosh, Soumya and Greenewald, Kristjan and Hoang, Nghia and Khazaeni, Yasaman},
  booktitle = {Proceedings of the 36th International Conference on Machine Learning (ICML)},
  year      = {2019},
  volume    = {97},
  series    = {Proceedings of Machine Learning Research},
  pages     = {7252--7261},
  publisher = {PMLR}
}

@misc{cortez2009winequality,
  author       = {Cortez, Pedro and Cerdeira, Ant{\'o}nio and Almeida, Francisco and Matos, Tiago and Reis, Jo{\~a}o},
  title        = {{Wine Quality} [Dataset]},
  howpublished = {UCI Machine Learning Repository},
  year         = {2009},
  doi          = {10.24432/C56S3T},
  note         = {\url{https://archive.ics.uci.edu/ml/datasets/wine+quality}}
}

@article{baptista2023federated,
  title        = {Federated Learning for Computer-Aided Diagnosis of Glaucoma Using Retinal Fundus Images},
  author       = {Baptista, Telmo and Soares, Carlos and Oliveira, Tiago and Soares, Filipe},
  journal      = {Applied Sciences},
  year         = {2023},
  volume       = {13},
  number       = {21},
  pages        = {11620},
  doi          = {10.3390/app132111620}
}

@article{XIA2024117151,
  title   = {Benchmarking deep models on retinal fundus disease diagnosis and a large-scale dataset},
  author  = {Xue Xia and Ying Li and Guobei Xiao and Kun Zhan and Jinhua Yan and Chao Cai and Yuming Fang and Guofu Huang},
  journal = {Signal Processing: Image Communication},
  volume  = {127},
  pages   = {117151},
  year    = {2024},
  issn    = {0923-5965},
  doi     = {10.1016/j.image.2024.117151},
  url     = {https://www.sciencedirect.com/science/article/pii/S0923596524000523},
  note    = {Dataset available at \url{https://drive.google.com/file/d/14haq2HifMv8rguGr8zUq8hM0TOblMzow/view}}
}

@ARTICLE{cartpole,
  author={Barto, Andrew G. and Sutton, Richard S. and Anderson, Charles W.},
  journal={IEEE Transactions on Systems, Man, and Cybernetics}, 
  title={Neuronlike adaptive elements that can solve difficult learning control problems}, 
  year={1983},
  volume={SMC-13},
  number={5},
  pages={834-846},
  doi={10.1109/TSMC.1983.6313077}}

@article{go2009twitter,
  title={Twitter Sentiment Classification using Distant Supervision},
  author={Go, Alec and Bhayani, Richa and Huang, Lei},
  journal={CS224N project report, Stanford},
  year={2009}
}

@misc{openai2025gpt41,
  author       = {OpenAI},
  title        = {GPT-4.1: Advancing Reasoning and Efficiency},
  year         = {2025},
  month        = {April},
  url          = {https://openai.com/index/gpt-4-1/},
  note         = {Accessed: 2025-09-14}
}

@article{mnih2013playing,
  title={Playing Atari with Deep Reinforcement Learning},
  author={Mnih, Volodymyr and Kavukcuoglu, Koray and Silver, David and Graves, Alex and Antonoglou, Ioannis and Wierstra, Daan and Riedmiller, Martin},
  journal={arXiv preprint arXiv:1312.5602},
  year={2013}
}

\end{document}